\documentclass[times,twocolumn,final]{elsarticle}

\usepackage{medima}
\usepackage{framed,multirow}
\usepackage{hyperref}
\usepackage{amssymb}
\usepackage{amsmath}
\usepackage{latexsym}
\usepackage{dsfont}
\usepackage{algorithm}
\usepackage{wrapfig}
\usepackage{algpseudocode}
\usepackage{caption}

\usepackage{url}
\usepackage{xcolor}
\usepackage{fancyhdr}

\journal{Arxiv}

\begin{document}

\verso{Weijie Chen \textit{et~al.}}

\begin{frontmatter}

\title{SASWISE-UE: Segmentation and Synthesis with Interpretable Scalable Ensembles for Uncertainty Estimation}
\author[1,2]{Weijie \snm{Chen}\corref{cor1}}
\author[1,2,3,4]{Alan \snm{McMillan}\corref{cor2}}
\cortext[cor2]{Corresponding author: 
  Email: \href{mailto:abmcmillan@wisc.edu}{wchen376@wisc.edu}}

\address[1]{Department of Electrical and Computer Engineering, University of Wisconsin-Madison, WI 53705, USA}
\address[2]{Department of Radiology, University of Wisconsin-Madison, WI 53705, USA}
\address[3]{Department of Medical Physics, University of Wisconsin-Madison, WI 53705, USA}
\address[4]{Department of Biomedical Engineering, University of Wisconsin-Madison, WI 53705, USA}

\availableonline{v1 on Nov 7, 2024}

\begin{abstract}
This paper introduces an efficient sub-model ensemble framework aimed at enhancing the interpretability of medical deep learning models, thus increasing their clinical applicability. By generating uncertainty maps, this framework enables end-users to evaluate the reliability of model outputs. We developed a strategy to develop diverse models from a single well-trained checkpoint, facilitating the training of a model family. This involves producing multiple outputs from a single input, fusing them into a final output, and estimating uncertainty based on output disagreements. Implemented using U-Net and UNETR models for segmentation and synthesis tasks, this approach was tested on CT body segmentation and MR-CT synthesis datasets. It achieved a mean Dice coefficient of 0.814 in segmentation and a Mean Absolute Error of 88.17 HU in synthesis, improved from 89.43 HU by pruning. Additionally, the framework was evaluated under corruption and undersampling, maintaining correlation between uncertainty and error, which highlights its robustness. These results suggest that the proposed approach not only maintains the performance of well-trained models but also enhances interpretability through effective uncertainty estimation, applicable to both convolutional and transformer models in a range of imaging tasks.
\end{abstract}

\begin{keyword}
\MSC[] 62P10 \sep 68T01\sep 68T37\sep 92C50
\KWD Uncertainty Estimation \sep Ensemble Learning \sep Image Segmentation\sep Image Synthesis \sep Medical Images
\end{keyword}

\end{frontmatter}


\section{Introduction}
Artificial intelligence (AI) has significantly influenced medical imaging, with applications ranging from image denoising (\cite{diwakarCTImageDenoising2020}) for enhanced clarity to classification (\cite{rajaBrainTumorClassification2020}), segmentation (\cite{isenseeNnUNetSelfconfiguringMethod2021}), and synthesis (\cite{huijbenGeneratingSyntheticComputed2024}), crucial for procedures such as attenuation and scatter correction in simultaneous PET/MRI (\cite{liuDeepLearningMR2018}). The evolution of network architectures has progressed from the foundational convolutional neural network, exemplified by U-Net (\cite{ronnebergerUnetConvolutionalNetworks2015}), to more sophisticated designs such as transformers (\cite{caoSwinUnetUnetlikePure2021}) and diffusion models (\cite{kazerouniDiffusionModelsMedical2022}). Beyond imaging, large language models have also been instrumental in extracting clinical concepts, medical relations, and assessing semantic textual similarities and natural language inference (\cite{yangLargeLanguageModel2022, abdarReviewUncertaintyQuantification2021}). Furthermore, multimodality models have demonstrated high accuracy in contexts such as the US Medical Licensing Exam (\cite{saabCapabilitiesGeminiModels2024}), indicating a growing trend of AI integration across various healthcare sectors.

In healthcare, the reliability and stability of models are of utmost importance due to their life-critical applications. Understanding how models perform with in-the-distribution data, out-of-distribution data, and in response to long-term dataset shifts is crucial. For online-running systems, model calibration is often employed to adjust for dataset shifts (\cite{huangTutorialCalibrationMeasurements2020}). In contrast, uncertainty estimation serves as a vital technique for interpreting results (\cite{abdarReviewUncertaintyQuantification2021}), offering insights into the confidence that can be placed in model outputs. This method not only enhances the interpretability of model predictions but also supports more informed decision-making in clinical settings.

Several methods exist for assessing uncertainty. One direct approach is to modify deep learning models to estimate uncertainty. For instance, (\cite{guoAnatomicMolecularMR2021}) extended a branch of a model's decoder specifically for this purpose, while (\cite{feiTwoPhaseMultiDoseLevelPET2024}) developed a separate network dedicated to uncertainty estimation. Additionally, various statistical methods have been developed to quantify uncertainty. The Dempster-Shafer framework, as detailed in (\cite{dempsterGeneralizationBayesianInference2008}) and applied in Evidence Deep Learning (\cite{sensoyEvidentialDeepLearning2018}), utilizes the Dirichlet distribution as a prior in a Bayesian model to express predictive confidence. Furthermore, the Kullback-Leibler divergence is used as a regularizer to discourage models from defaulting to uncertain predictions. These methodologies represent significant strides in the effort to quantify and manage uncertainty in medical applications.

Another prominent method for handling uncertainty is ensemble learning. This approach involves creating variations in data (\cite{ashukhaPitfallsInDomainUncertainty2021, karimiAccurateRobustDeep2019}) model hyper-parameters (\cite{lakshminarayananSimpleScalablePredictive2017}), or algorithms (\cite{delarosaRobustEnsembleAlgorithm2024}). Additionally, Bayesian networks represent a well-established technique for uncertainty estimation. Probabilistic models such as Bayesian networks (\cite{galDropoutBayesianApproximation2015}) use neural networks with dropout layers (\cite{zhaoBayesianConvolutionalNeural2018}) as a means to approximate the Monte Carlo random process. Dropout can be implemented pixel-wise or channel-wise (\cite{houWeightedChannelDropout2019}), effectively creating ensembles of sub-models during model evaluation. Furthermore, regularized dropout (\cite{liangRDropRegularizedDropout2021}) improves stability and convergence speed by minimizing performance discrepancies among sampled sub-models. Taking ensemble methods further, an efficient ensemble approach (\cite{leeEfficientEnsembleModel2020}) trains a single model with consistent hyper-parameters and dataset but introduces variability by randomly replacing layers, generating an exponential number of models with only a linear increase in training time, thus enhancing inter-model uncertainty estimation.
\begin{figure*}[!t]
\centering
\includegraphics[width=0.85\linewidth]{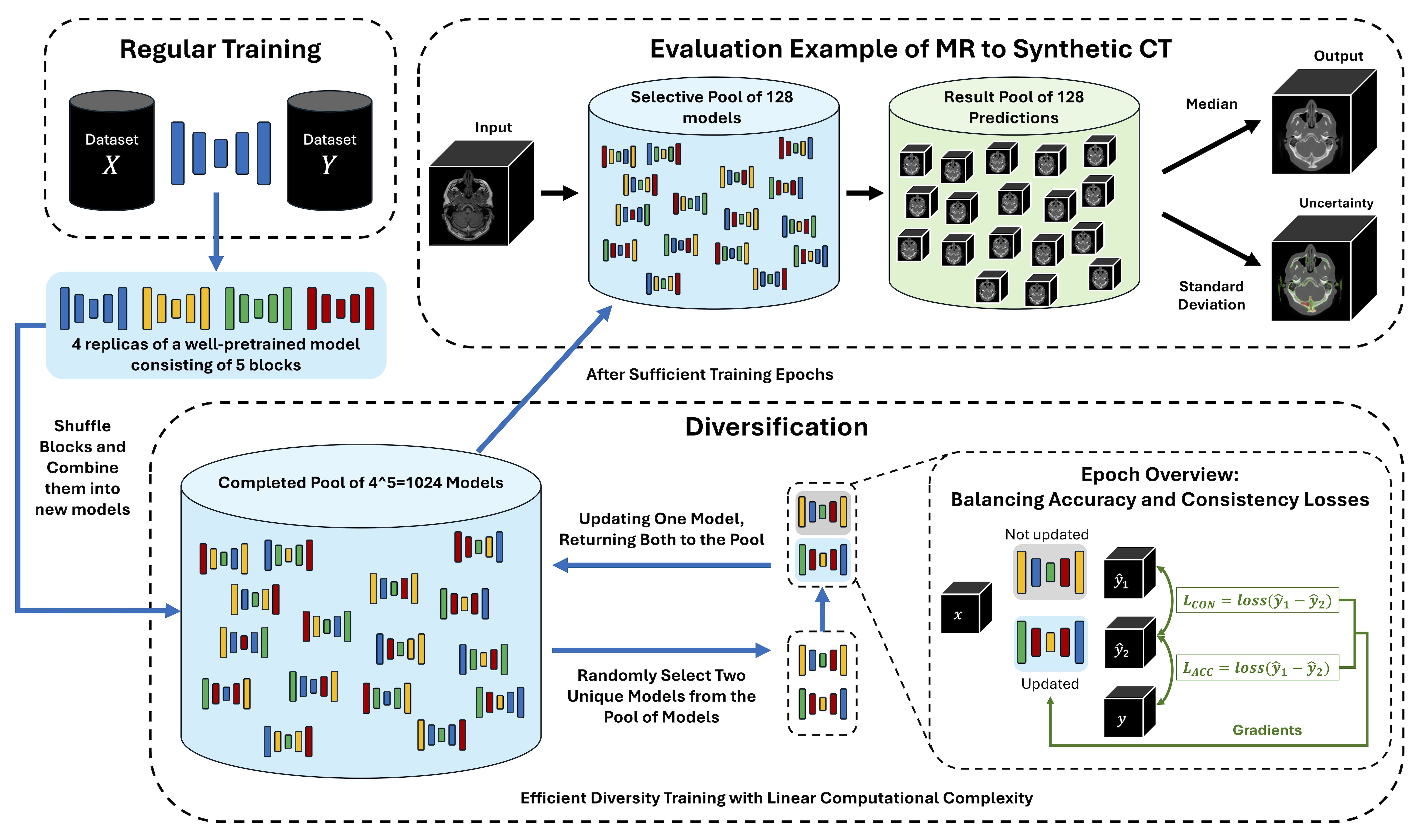}
\caption{The SASWISE pipeline efficiently estimates uncertainty while maintaining or enhancing pre-trained model performance. It begins by training a supervised model to convergence, followed by creating multiple candidate blocks in each block position. These blocks are shuffled and recombined into new models. In the diversification stage, two unique models are selected from the pool and trained on the same data sample. This stage involves calculating and utilizing the accuracy loss between the model being updated and the ground truth, along with the consistency loss between the two models, to only update the model being refined. After enough diversification training epochs, the best models from the partial or complete model pool are used to generate results from a single input. The final results for tasks with continuous or discrete data types are determined using median or majority voting methods, respectively, with uncertainty maps produced using standard deviation or majority ratio.}
\label{fig_GA}
\end{figure*}

In this study, we introduce an efficient ensemble method called SASWISE. An example is demonstrated on Figure \ref{fig_GA}. The SASWISE pipeline is designed to efficiently estimate uncertainty while preserving or enhancing the performance of a pre-trained model. Initially, a supervised model is trained to convergence. Subsequently, multiple copies of each block are created, resulting in various candidate blocks for each block position. These candidate blocks are then shuffled and recombined into new models. During the diversification stage, two unique models are randomly selected from the pool and trained on the same data sample. The accuracy loss between the model being updated and the ground truth, along with the consistency loss between the two selected models, are calculated. These losses are utilized solely to update the model being refined. After empirically enough diversification training epochs, either the best selective model pool or the complete model pool is used to generate a results pool from a single input data point. For continuous or discrete data type tasks, the final result is produced using median or majority vote methods, respectively. Additionally, the uncertainty map is generated using either standard deviation or majority ratio. This methodology not only maintains but potentially improves the performance of the given pre-trained model, while utilizing resources and time efficiently.
\section{Materials and methods}

\subsection{Problem setup}
In the traditional supervised learning model, a dataset $D$ consisting of $N$ paired samples, represented as $D={(x_1,y_1 ),(x_2,y_2 ),…,(x_N,y_N )}$ is used. Each sample pair $(x_i,y_i )$ randomly chosen based on the distribution of $D$. A neural network with parameters $\theta$, is utilized to approximate the function $f_\theta$, which predicts the output $y_i$ from the input $x_i$, formally denoted as
$$f_\theta(x_i)=\hat{y_i}$$

To modularize this model, the network can be segmented into two sequential components $f_{\theta_1}$ and $f_{\theta_2}$, allowing the input to be processed in a cascaded fashion to produce the output:
$$f_{\theta_2}(f{_\theta}_1(x_i))=\hat{y_i}$$

Furthermore, there can be homogeneous candidate blocks in each block position. For instance, $f_{\theta_1^1}$ and $f_{\theta_1^2}$, serve as the first and second candidate block for the initial block position $f_{\theta_1}$, respectively, and similarly,  $f_{\theta_2^1}$ and $f_{\theta_2^2}$ for the subsequent block position $f_{\theta_2}$. This framework generates four distinct predictive outcomes:
$$f_{\theta_2^1}(f_{\theta_1^1}(x_i))=\hat{y_i}^{11}$$
$$f_{\theta_2^1}(f_{\theta_1^1}(x_i))=\hat{y_i}^{12}$$
$$f_{\theta_2^1}(f_{\theta_1^2}(x_i))=\hat{y_i}^{21}$$
$$f_{\theta_2^2}(f_{\theta_1^2}(x_i))=\hat{y_i}^{22}$$

Expanding this concept, the network partitioning can be generalized to $P$ positions, each with $A_i$ candidates $(1\leq i\leq P)$. By selecting a specific candidate $k_i$ $(1\leq k_i\leq A_i)$ block in each position, the predictive model can be represented as:
$$f_{\theta_P^{k_P}}(\dots f_{\theta_2^{k_2}}(f_{\theta_1^{k_1}}(x_i)))=\hat{y_i}^{k_1k_2\dots k_P}$$

This modular approach allows us to delineate a combination, or a "path" $\omega$ through the network, using a sequence of selected candidates $\omega=k_1k_2\dots k_P$, which signifies the route the input data takes through various blocks before culminating in a prediction. This methodology not only facilitates a flexible architecture but also provides a structured mechanism to explore different configurations for optimizing performance.

\subsection{Model setup}
\begin{figure*}[!t]
\centering
\includegraphics[width=0.85\linewidth]{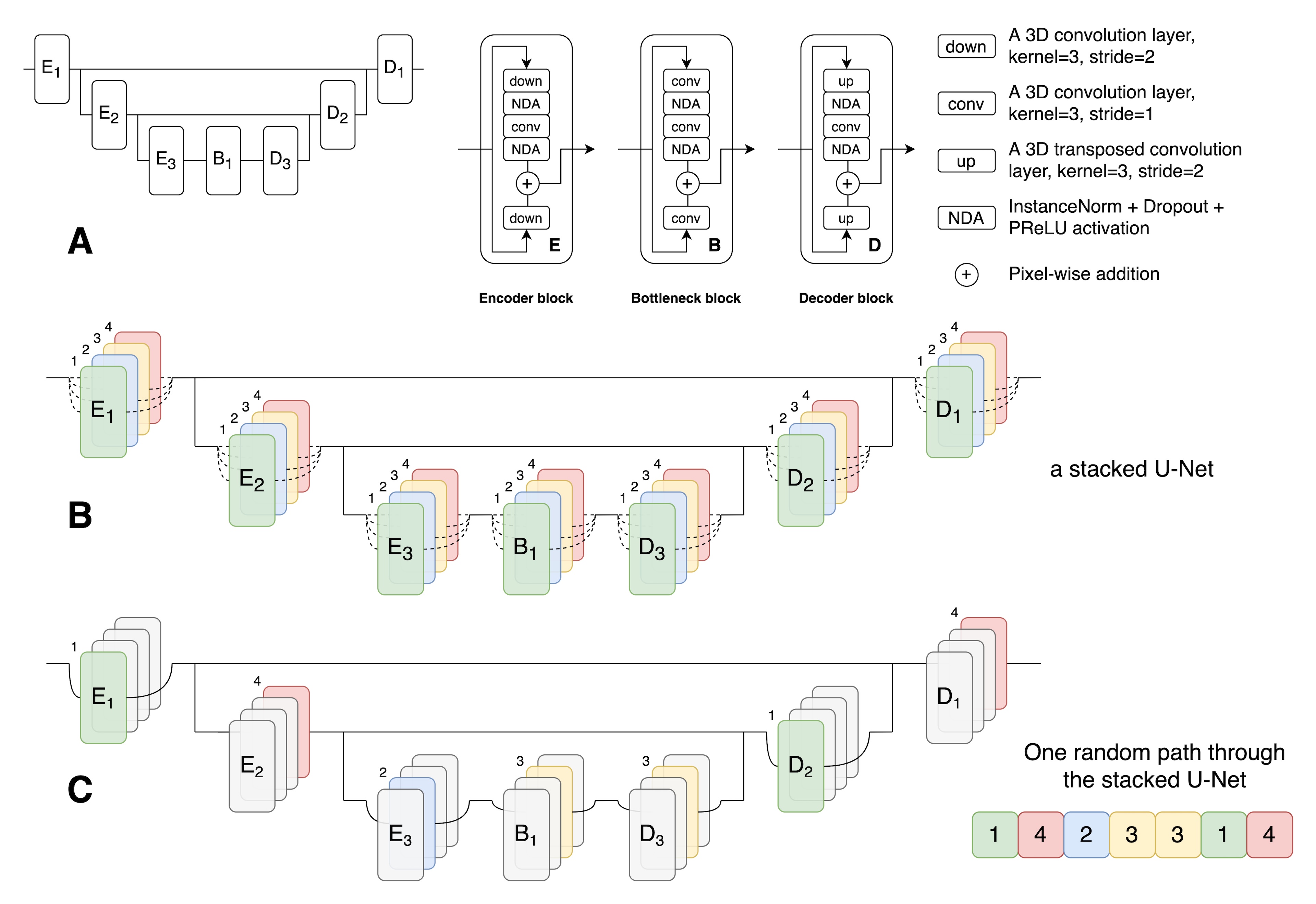}
\caption{Schematic illustration of the SASWISE approach using a stacked U-Net. (A) The standard U-Net after full training epochs, serving as the template. (B) The template is replicated by cloning weights of each block and stacking them. (C) During a single data flow, one random path is active, represented by the colored bricks. E for encoder block, B for bottleneck block, and D for decoder block.}
\label{fig_PD}
\end{figure*}

Initially, we implement the SASWISE system using the U-Net model (\cite{ronnebergerUnetConvolutionalNetworks2015}), a widely-used encoder-decoder model. Figure \ref{fig_PD}A depicts a U-Net implementation created in Project MONAI (\cite{cardosoMONAIOpensourceFramework2022}), comprising encoder blocks, bottleneck blocks, and decoder blocks. Each block contains alternating stacked layers of convolution/transposed-convolution, normalization, dropout (deactivated by default), PReLU activation, and residual paths. Additionally, we employ the UNETR model (\cite{hatamizadehUNETRTransformers3D2022}), which substitutes the encoder and bottleneck blocks in U-Net with a transformer block.

After the U-Net has completed its designated training epochs, we construct a stacked version by duplicating each block with identical weights to create multiple replicas (four as depicted in Figure \ref{fig_PD}B). These replicas inherit weights from the initially trained U-Net, establishing a foundation for diversified blocks through subsequent training. This stacking arrangement can be generalized across different levels within the model, with each layer initiating from a copied candidate block—allowing for subsequent divergence in function.

During each network pass, only one candidate block per position is activated, as shown in Figure \ref{fig_PD}C, which displays a random activation path through the stacked U-Net, represented by $\omega=[1,4,2,3,3,1,4]$. Active blocks are highlighted, whereas inactive blocks remain gray, emphasizing the dynamic utilization of block candidates within the network.

\subsection{Training and evaluation}
\begin{figure*}[!t]
\centering
\includegraphics[width=0.85\linewidth]{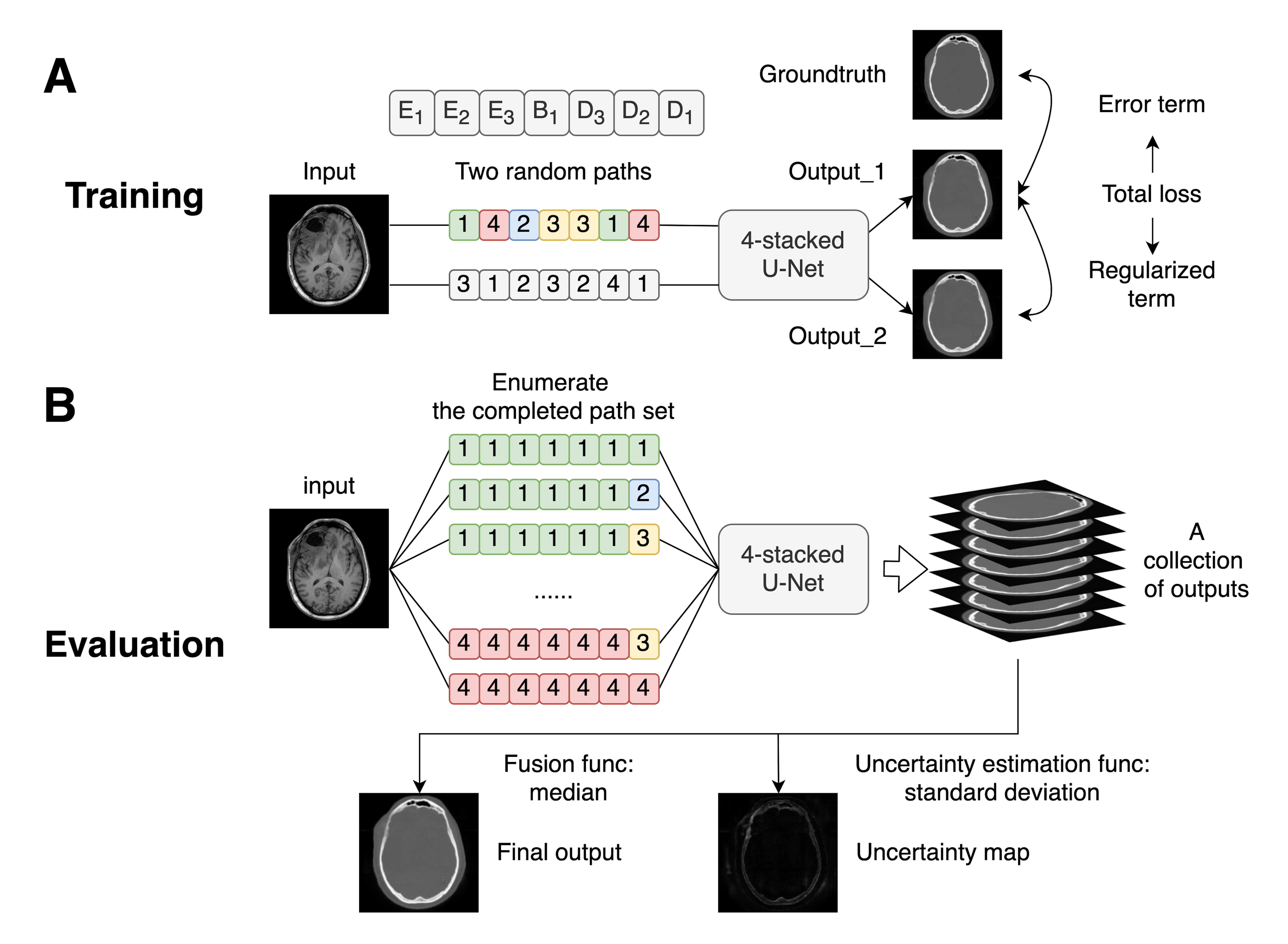}
\caption{Training and Evaluation Protocol for an Image Synthesis Application. (A) During the training phase, the same input is processed through two distinct paths, optimizing the overall loss which includes the error term (comparing the prediction with the ground truth for accuracy) and a regularization term (assessing consistency between the two predictions). (B) In the evaluation phase, all potential paths or a selected subset of models from the completed path set are considered, and the final prediction is derived using a fusion function. Additionally, uncertainty is estimated using a dedicated uncertainty estimation function.}
\label{fig_TV}
\end{figure*}

In the training phase (Figure \ref{fig_TV}A), it is essential to fully train the template model before introducing candidate blocks. Initiating diversification with the unconverged checkpoint can lead to inconsistent optimization across epochs, resulting in fluctuating performance and potential convergence challenges within the loss function landscape. For each training epoch of the stacked U-Net, two paths, $\omega$ and $\omega'$, are randomly selected for each input batch, yielding two separate predictions, $\hat{y}^\omega$ and $\hat{y}^{\omega'}$. The loss function is then augmented with a regularization
\noindent
\begin{minipage}{\linewidth}
\begin{algorithm}[H]
\caption{The training for our proposed SASWISE.}\label{alg:train}
\begin{algorithmic}[1]
\State Prepare Dataset $D$ containing $N$ samples and model $f_{\theta}$
\Repeat
\State Draw an random sample $(x_i, y_i)$ to train the model $f_{\theta}$
\Until{$f_{\theta}$ converges}
\State Divide $f_{\theta}$ into $P$ segments
\State Duplicate $A_i$ candidate blocks for each position $P_i$
\State Build a path pool $\Omega=\{\omega=(k_1,k_2,k_3\dots,k_P)\,|\, 1\leq k_i\leq A_i\}$
\State Build a model pool, $f_\theta(;\omega\in\Omega)=f_{\theta_P^{k_P}}(\dots f_{\theta_2^{k_2}}(f_{\theta_1^{k_1}}(x_i)))$
\Repeat
\State Draw an random sample $(x_i, y_i)$
\State Sample two path $\omega$ and $\omega\prime$ from $\Omega$
\State Compute $\hat{y}_\omega=f_\theta(x_i;\omega)$ and $\hat{y}_{\omega\prime}=f_\theta(x_i;\omega\prime)$
\State $Loss=loss\_fn(y_i, \hat{y}_\omega)+\alpha\,loss\_fn(\hat{y}_\omega, \hat{y}_{\omega\prime})$
\State Update blocks in path $\omega=(k_1,k_2,k_3\dots,k_P)$
\Until{Diversification ends}
\end{algorithmic}
\end{algorithm}
\end{minipage}
 term that addresses the discrepancy between these two predictions, in addition to the standard error term that measures the deviation between the ground truth $y$ and the prediction $\hat{y}^\omega$. 
In this manuscript, two losses are balanced using the coefficient $\alpha$.For simplicity, $\alpha$ is set to 1, but it can be adjusted according to user requirements. During back-propagation, only the candidate blocks 
along the active path, $\omega$, responsible for the prediction $\hat{y}^\omega$, are updated. This focused updating approach helps in stabilizing the learning process and refining the model's performance by iteratively adjusting only the actively involved blocks, illustrated in Algorithm \ref{alg:train}

\noindent
\begin{minipage}{\linewidth}
\begin{algorithm}[H]
\caption{The evaluation for our proposed SASWISE.}\label{alg:eval}
\begin{algorithmic}[1]
\State Given the testing dataset $T$
\State Given the complete, random or selected path pool $\Omega$ and corresponding model pool $f_\theta(;\omega\in\Omega)$
\State Given fusion function $\mu$ and uncertainty estimation function $\sigma$
\State \textbf{For $(x, y) \in T$} 
\State \hspace{1.25em} Initialize the result pool $\hat{Y}_{\Omega}=\{\}$
\State \hspace{1.5em} \textbf{For $\omega_i\in \Omega$}
\State \hspace{1.25em} \hspace{1.25em} Append $\hat{y}_{i}=f_\theta(x;\omega_i)$ to $\hat{Y}_{\Omega}$
\State \hspace{1.25em} \textbf{End}
\State \hspace{1.25em} Compute final prediction $\hat{y}_\Omega=\mu(\hat{Y}_{\Omega})$
\State \hspace{1.25em} Compute uncertainty estimation
$\tilde{y}_\Omega=\sigma(\hat{Y}_{\Omega})$
\State \textbf{End}
\end{algorithmic}
\end{algorithm}
\end{minipage}
\\

In the evaluation stage, as depicted in Figure \ref{fig_TV}B, we have the option to either enumerate the entire set of completed paths, or to prune the model pool to select a subset of models that demonstrate superior performance or faster inference. From this result pool $\hat{Y}_\Omega$, we derive the collection of outputs, $\hat{y}_\Omega$. Utilizing this collection, a fusion function is employed to synthesize the final prediction, and an uncertainty estimation function is applied to gauge the uncertainty associated with that prediction, illustrated in Algorithm \ref{alg:eval}.
\[
\hat{y}_{discrete} = \operatorname{mode}(\hat{y}_\Omega) = \operatorname{arg\,max}_{y} |\{y_i \in \hat{y}_\Omega : y_i = y\}|
\]

In contrast, for continuous variables, the \textbf{median} of the outputs is used to achieve a central tendency, as below where $\hat{y}\in Y_{\Omega}$ and $Y_{\Omega}$ is sorted.
\[
\hat{y}_{continuous} = \operatorname{median} = \begin{cases} 
\hat{y}_{\frac{n+1}{2}} & \text{if } n \text{ is odd}\\
\frac{1}{2}(\hat{y}_{\frac{n}{2}} + \hat{y}_{\frac{n}{2} + 1}) & \text{if } n \text{ is even}
\end{cases}
\]

To quantify uncertainty, \textbf{the degree of disagreement in the majority vote} for discrete variables is calculated by taking the ratio of the most common choice and subtracting this value from one, serving as a fundamental measure of variation.

\[
\delta_{\{\hat{y}=\hat{y}_{discrete}\}} = \begin{cases} 
1 & \hat{y}=\hat{y}_{discrete} \\
0 & \text{otherwise}
\end{cases}
\]

\[
\text{Uncertainty}_{ discrete} = 1-\frac{1}{N}\sum_{i=1}^N \delta_{\{\hat{y}=\hat{y}_{discrete}\}}
\]

For continuous variables, the \textbf{standard deviation} of the outputs is used to provide a measure of dispersion, offering insights into the reliability of the predictions. 

\[
\text{Uncertainty}_{ continuous} = \sqrt{\frac{1}{n-1} \sum_{i=1}^N (\hat{y}_i - \bar{y})^2}
\]


\subsection{Pruning}
\begin{figure*}[!t]
\centering
\includegraphics[width=0.75\linewidth]{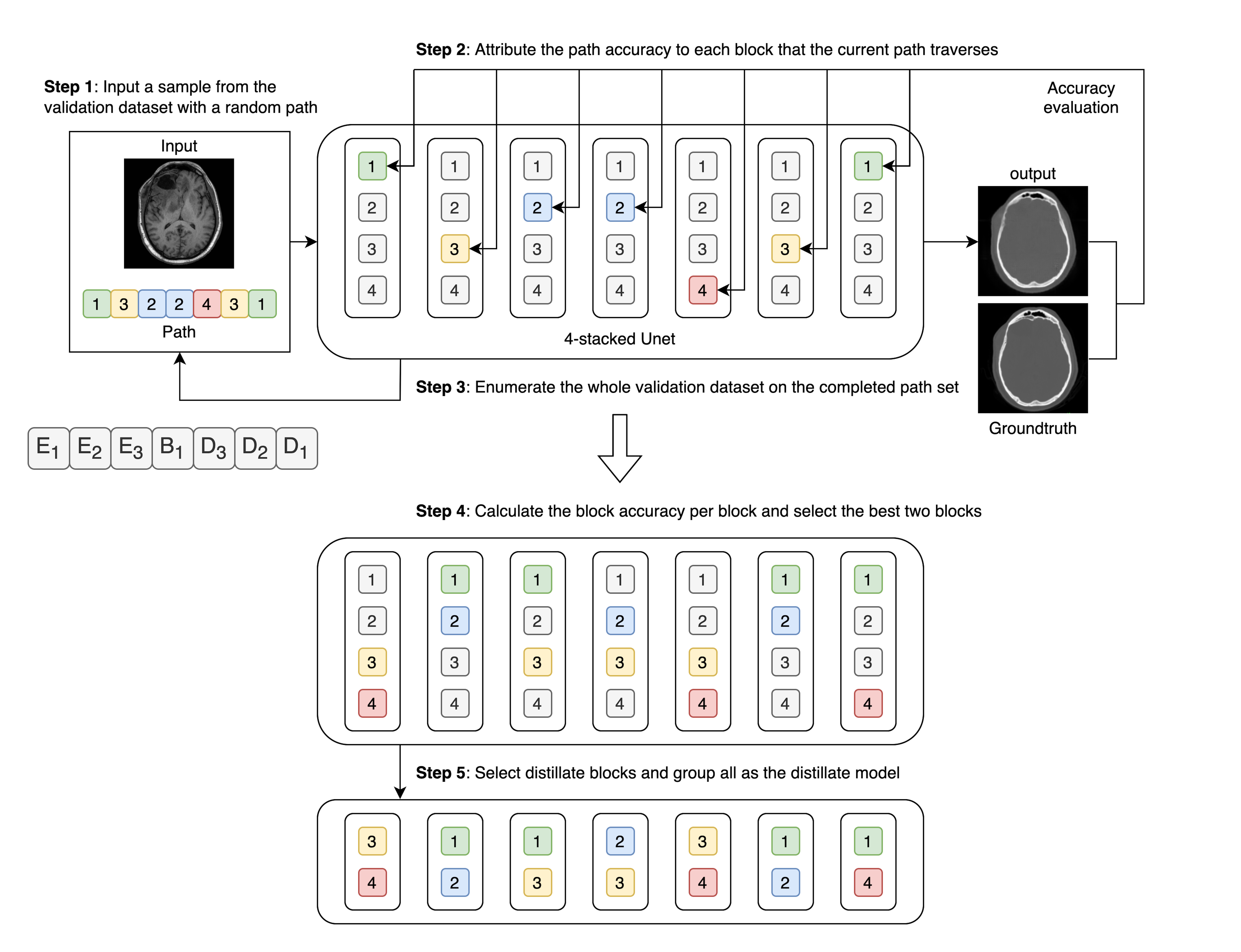}
\caption{Diagram of the Model Pruning Process. Step 1: Review a validation dataset using various randomly selected paths. Step 2: Determine the accuracy for each path and allocate this accuracy to the blocks traversed. Step 3: Continue this procedure until the entire validation dataset has been evaluated across the available paths. Step 4: Compute the average accuracy for each block and retain the blocks with the highest accuracy in each stack. Step 5: Construct a refined model using the selected high-performing blocks.}
\label{fig_PP}
\end{figure*}
Given the scenario where the model is segmented into seven blocks, each having four candidate blocks, the total number of possible combinations is 
$4^7=16384$. Exploring all possible paths is impractical and time-consuming. We have two strategies to manage this complexity: we can either randomly select a number of paths for evaluation, or we can reduce the number of candidate blocks in each position from four to two based on preliminary assessments.

As illustrated in Figure \ref{fig_PP}, when a prediction is made through a specific path, the error between the prediction and the ground truth is assigned to all candidate blocks along that path. By evaluating the validation dataset, we can determine the performance of each candidate block based on the mean error across all paths that traverse these blocks. Subsequently, candidate blocks that exhibit inferior performance relative to others at the same position are pruned. This pruning process results in a streamlined model pool, reducing complexity while potentially maintaining or even enhancing model performance, illustrated in Algorithm \ref{alg:prune}
\noindent
\begin{minipage}{\linewidth}
\begin{algorithm}[H]
\caption{The pruning for our proposed SASWISE.}\label{alg:prune}
\begin{algorithmic}[1]
\State Given the validation dataset $V={(x_i,y_i)}$
\State Given the path pool $\Omega=\{\omega=(k_1,k_2,k_3\dots,k_P)\}$
\State Given the model pool, $f_\theta(;\omega\in\Omega)=f_{\theta_P^{k_P}}(\dots f_{\theta_2^{k_2}}(f_{\theta_1^{k_1}}(x_i)))$
\State Initialize the metric pool $M(j,k)$, where store the performance of the $k$th candidate block in the $j$th position.
\State Given the performance metric function $\psi$
\State \textbf{For} Sample $(x_i, y_i) \in T$, $\omega=(k_1,k_2,k_3\dots,k_P) \in \Omega$
\State \hspace{1.25em} Compute $\hat{y_i}=f_\theta(x_i;\omega)$
\State \hspace{1.25em} Compute performance $\psi(y_i, \hat{y_i})$
\State \hspace{1.25em} Append $\psi(y_i, \hat{y_i})$ to $M(j, k_j)$, where $j=1,2,\dots,P$
\State \textbf{End}
\State \textbf{For} position $j$=1 to $P$
\State \hspace{1.25em} Compute mean $\bar{\psi}_{jk}=\frac{1}{\lvert M(j, k) \rvert} \sum m$, where $m\in M(j, k) $
\State \hspace{1.25em} Sort candidate blocks in position $j$ by \( \bar{\psi}_{jk} \) descendingly
\State \hspace{1.25em} Select top-performing blocks
\State \textbf{End}
\State Create new path pool $\Omega_{\text{new}}$
\State Create new model pool \( f_{\theta_{\text{new}}} \) based on \( \Omega_{\text{new}} \)
\end{algorithmic}
\end{algorithm}
\end{minipage}

\subsection{Data preparation}
For the segmentation task, we employed the BTCV dataset (\cite{gibsonAutomaticMultiOrganSegmentation2018}), containing 15/9/6 patient-level cases for training/validation/testing and 13 distinct organs for segmentation. For the synthesis task, generating synthetic CT from MR images for attenuation correction in PET/MR or radiation therapy, we used the MR-CT dataset obtained from our institution with a waiver of consent (\cite{estakhrajiEffectTrainingDatabase2023a}). Patient selection criteria included having both CT and MR head imaging within 48 hours, with MR imaging performed at 1.5T (MR450w or HDxt, GE Healthcare) using a specific pulse sequence: 3D T1-weighted inversion prepared gradient-recalled echo, echo time = 3.5 ms, inversion time = 450 ms, repetition time = 9.5 ms, in-plane resolution: 0.98 × 0.98 mm, slice thickness = 2.4–3.0 mm. CT exams were performed on one of three scanners (Optima CT 660, Discovery CT750HD, or Revolution GSI, GE Healthcare) with specific acquisition and reconstruction parameters: 0.45 × 0.45 mm transaxial resolution, 1.25 or 2.5 mm slice thickness, 120 kVp, automatic exposure control with noise index of 2.8–12.4, and helical pitch of 0.53. The dataset underwent intensity normalization. The CT dataset was clipped between -1000 and 3000 and normalized to the [0,1] range using Max-min normalization. The MR dataset was clipped between 0 and 10000 and similarly normalized to the [0,1] range using Max-min normalization. Augmentation techniques, such as cropping, flipping, rotation, and intensity shift, were applied to expand the training data population. A total of 660 cases were normalized and randomly divided 5:3:2 for training, validation, and testing. MRI and CT images were spatially registered using the ANTsPy package with Symmetric normalization and resampled to a 1x1x1 mm voxel size.
\subsection{Implementation}
We utilized U-Net and UNETR models, denoted by E, B, D for encoders, bottleneck, and decoders, and T and D for transformer blocks and decoders, respectively. We experimented with various U-Net (E2B2D2, E4B4D4, E1B4D4, E4B4D1) and UNETR (T2D2, T1D4, T4D1) models for segmentation, while employing E2B2D2 for synthesis and three distilled synthesis models (E1B4D4, E4B4D1, E4B4D4). All models used 96x96x96 voxel patches, implemented using MONAI on an Nvidia V100 GPU. To ensure a fair comparison, each SASWISE ensemble evaluated all possible paths during its assessment. For instance, the E2B2D2 configuration, which has $2^7=128$ paths, and the E2B2D1 configuration with $2^6=64$ paths, sampled every path available. In cases where the total number of paths exceeded 128, a selection of 128 paths was made randomly and without repetition. For example, the E4B4D4 configuration, with its $4^7=16384$ paths, and the E4B4D1 configuration with $4^4=256$ paths, both utilized a distinct, randomly chosen set of 128 paths for evaluation. For the synthesis task, we examined pre-training effects, comparing an SASWISE model trained for 50 epochs ($\text{E2B2D2}_{\text{EarlyStop}}$) to one trained for 200 epochs ($\text{E2B2D2}_{\text{FullTrain}}$). We employed the SciPy package to quantitatively compare output models using specific metrics. For segmentation, we used the Dice coefficient per organ/region, while for synthesis, we assessed mRMSE, MAE, SSIM, PSNR, and acutance. Acutance, correlating with image sharpness, represents the mean absolute value of voxel-wise gradients, with higher values indicating sharper edges. The body contours are acquired by extracting the top 95\% of MR values and are further refined to ensure continuity and completeness of the contour shapes. We compared all results to a baseline U-Net using the Wilcoxon signed-rank test, applying a Bonferroni correction to adjust for multiple comparisons.
\subsection{Other dropout models}
We compared alternative ensemble model approaches, including basic ensembles and Monte Carlo dropout methods. The basic ensemble approach trained separate models with distinct random seeds, increasing training time proportionally to the number of models. We developed 32 unique models, requiring 32 times the resources of a single model, without pre-training. Additionally, we implemented voxel-wise and channel-wise Monte Carlo dropout. Voxel-wise dropout aligns with conventional node-based dropout methods. We also utilized channel dropout and weighted channel dropout. Channel dropout randomly discards an entire channel, while weighted channel dropout (\cite{houWeightedChannelDropout2019}) trains twice as many channels and scores them based on average intensity. Channels are randomly selected with probabilities corresponding to their scores, favoring stronger signals. We also included a method that shuffles channels for comparison. The Monte Carlo voxel/channel dropout rate and weighted channel dropout rate were set at 0.25
\subsection{Correlation Analysis}
This section explores how uncertainty aids the task by inspecting the error through the uncertainty map. To analyze the relationship between error and uncertainty, we calculate various correlations between these two metrics. Visually, for segmentation tasks, we delineate a boundary line where uncertainty exceeds a specific threshold, highlighting potential boundary contours. In image synthesis, the uncertainty map is segmented into four distinct regions—transparent, green, yellow, and red—each indicating a different level of uncertainty. This color-coded guide directs experts' attention to scrutinize the prediction quality more effectively.

At both voxel and case levels, the correlation between error and uncertainty is quantitatively assessed. Additionally, we compute the Pearson coefficient to determine the linear correlation between these variables, providing insights into how closely uncertainty can predict error. This approach not only enhances understanding of model reliability but also aids in refining model accuracy through targeted adjustments based on identified uncertainties.

We compute both the Dice coefficient and IoU (Intersection over Union) scores for regions with lower and higher errors, as well as for areas with lower and higher uncertainty. These metrics help demonstrate the overlap and assist users in identifying error-prone regions via the uncertainty map.
\subsection{MRI Corruption}
In this section, we incrementally corrupt the input data (MR images) to observe changes in error and uncertainty corresponding to varying levels of corruption. Four types of noise are introduced: Gaussian noise, Rician noise, Rayleigh noise, and Salt \& Pepper noise. Gaussian noise, a common image noise, affects the image uniformly. Rician noise arises because MR images represent the magnitude of complex numbers; when independent Gaussian noise affects both the real and imaginary parts, the resulting magnitude contains Rician noise. Rayleigh noise typically appears in systems where the overall magnitude of a vector is measured, composed of orthogonal components each with Gaussian distribution. Salt \& Pepper noise randomly assigns pixels to either the minimum or maximum value, though it is less common in MR images.

Additionally, two under-sampling methods are introduced to mimic real-life practices that reduce MR scanning time by under-sampling the K-space (frequency space). The Radial and Spiral trajectories are employed in each axial slice of MR images, resulting in lower-quality images due to fewer points sampled in K-space. The levels of corruption for all six types are progressively increased to assess how the model responds and how the uncertainty-error correlation evolves under these conditions.
\section{Results}
\subsection{Segmentation}
\begin{figure*}[!t]
\centering
\includegraphics[width=0.75\linewidth]{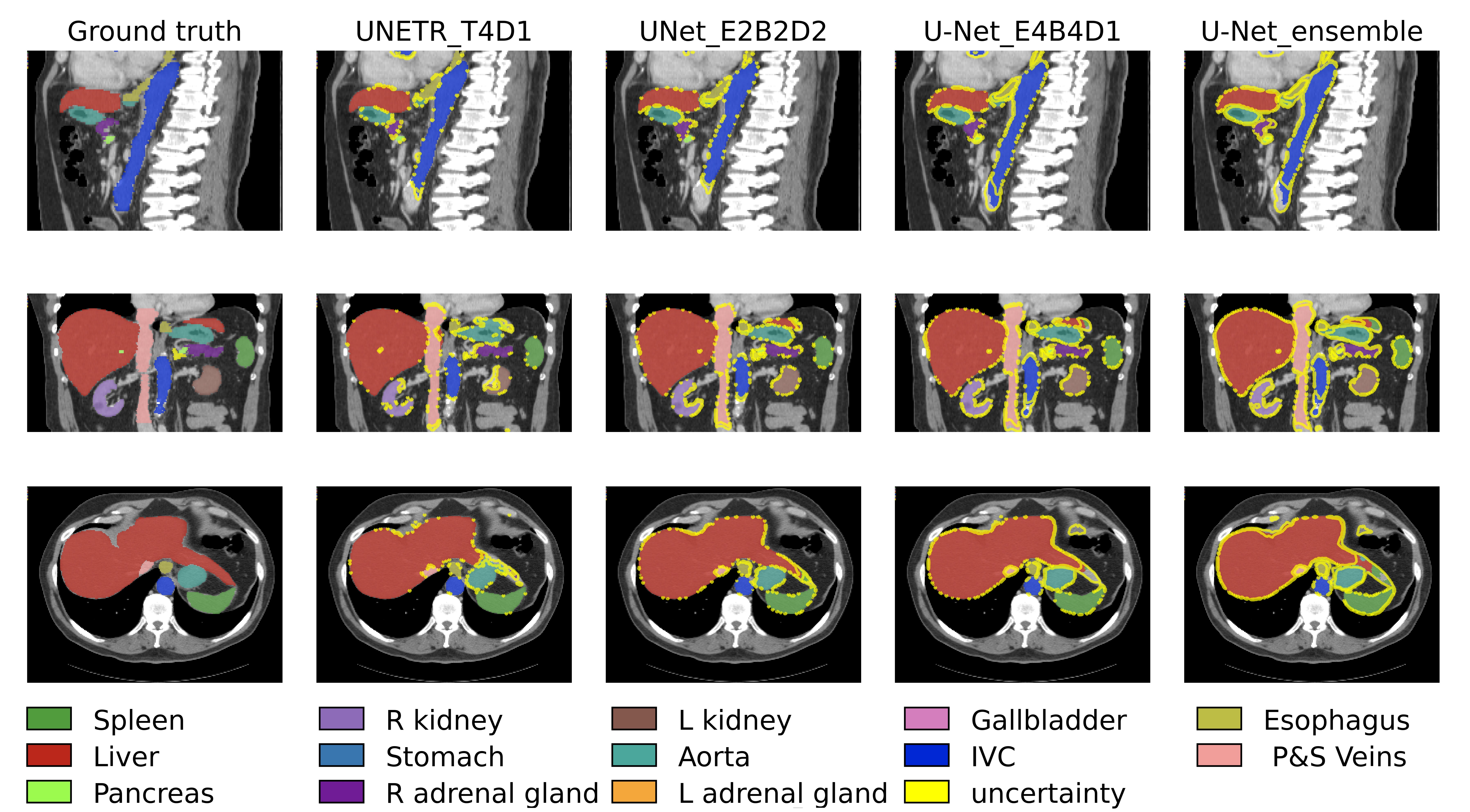}
\caption{Visual analysis of segmentation models. Column 1 displays the ground truth. Column 2, ``UNETR\_T4D1,'' combines four transformer decoders with a U-Net decoder. Column 3, ``U-Net\_E2B2D2,'' integrates two U-Net encoders, bottleneck structures, and decoders. Column 4, ``U-Net\_E4B4D1'' employs four U-Net encoders and bottleneck structures, ending with one decoder. Column 5, ``U-Net\_ensemble,'' compiles outputs from thirty-two distinct U-Nets. Yellow lines indicate segmentation disagreement ratios among predictions greater than \(7\%\) (\(\frac{1}{14}\), for 14 classes).}
\label{fig_ReSeg}
\end{figure*}
Table 1 in the segmentation experiment shows a performance comparison between U-Net and UNETR models using various configurations. Some models struggled with segmenting regions, indicating limitations in handling intricate details, especially in a large ratio of dropout models. However, the proposed SASWISE technique consistently outperformed the baseline model for both U-Net and UNETR models, demonstrating that the SASWISE approach effectively improves segmentation performance by leveraging ensemble learning and stacked model structures. Figure 4 supports this conclusion, displaying example segmentation outputs and uncertainty maps with superior results achieved by the SASWISE technique.
\begin{figure*}[!t]
\centering
\captionof{table}{Dice coefficients for U-Net and UNETR models. We present naive models, dropout models, basic ensemble models, and the proposed SASWISE approach for both U-Net and UNETR models. Bolded numbers represent the alpha = 0.05 Bonferroni-corrected significance level of the Wilcoxon signed-rank test. Empty slots indicate segmentation failure for specific regions. IVC = Inferior vena cava, PVSV = Portal vein and splenic vein, MC = Monte-Carlo. }
\label{table_ReSeg}
\includegraphics[width=\linewidth]{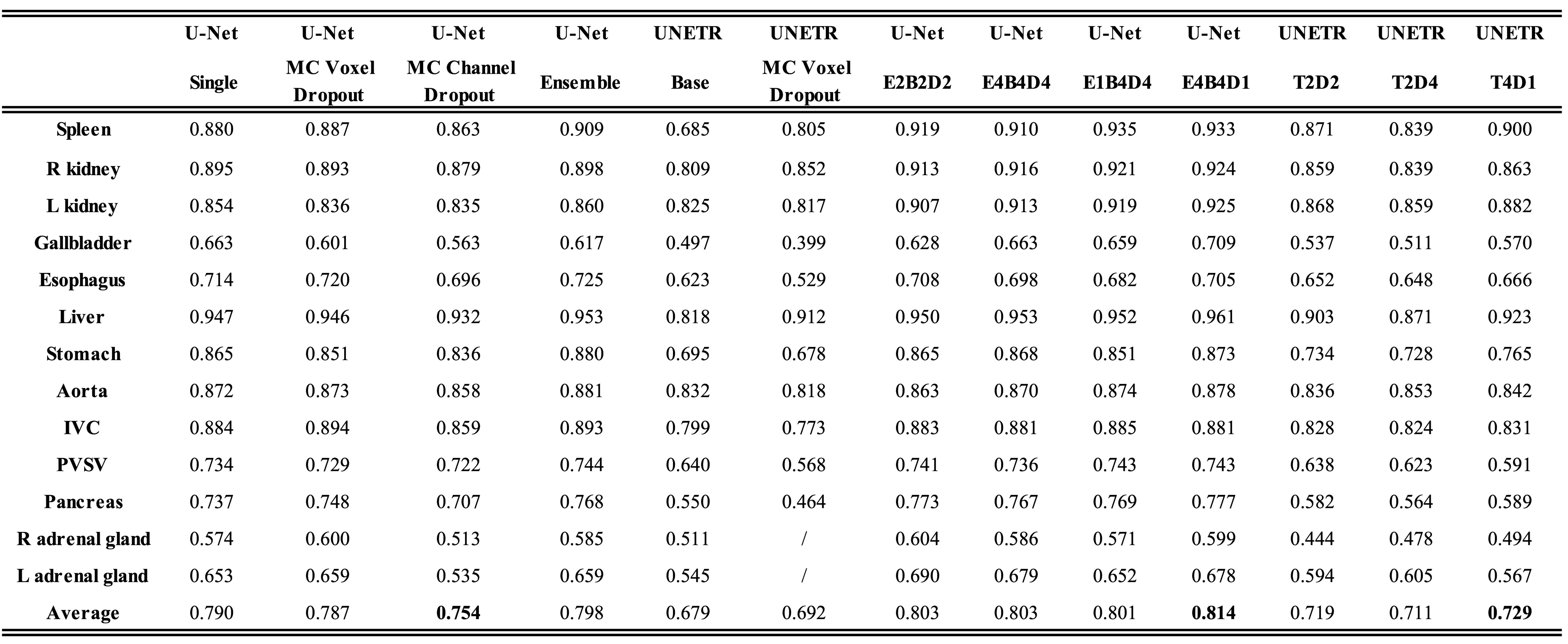}
\end{figure*}
\subsection{Synthesis}
\begin{figure*}[!t]
\centering
\includegraphics[width=0.75\linewidth]{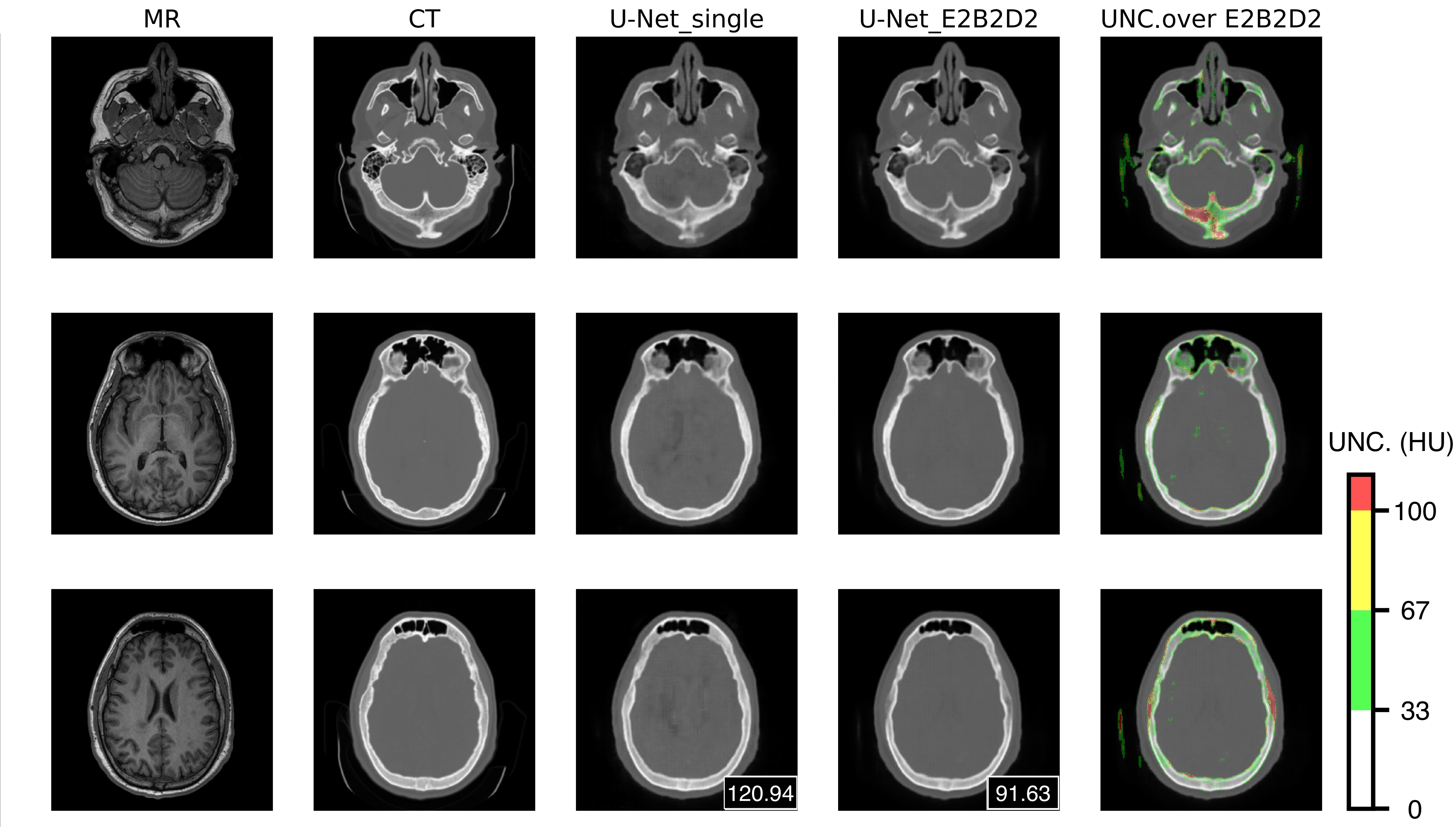}
\caption{Visual analysis of MR-to-CT synthesis. MR-to-CT Image Synthesis: Showcasing three axial views, the third column uses a single U-Net model, and the fourth column features an advanced U-Net variant with dual encoders, bottlenecks, and decoders. White numbers indicate the Mean Absolute Error (MAE) in Hounsfield Units (HU), quantifying synthesis accuracy. In the fifth column's uncertainty maps, transparency indicates standard deviations below 33 HU for high confidence, with increasing levels of uncertainty marked in green (33-67), yellow (67-100), and red (over 100).}
\label{fig_ReSyn}
\end{figure*}
Table 2 summarizes the synthesis experiment results, showing that MC dropout approaches, particularly channel dropout, underperformed compared to the baseline model, indicating their limitations in this task. However, the proposed SASWISE models demonstrated better performance than the baseline model, proving the efficacy of the SASWISE technique in synthesis tasks. Interestingly, the SASWISE model trained for 50 epochs ($\text{E2B2D2}_{\text{EarlyStop}}$) showed slightly inferior performance compared to the one trained for 200 epochs ($\text{E2B2D2}_{\text{FullTrain}}$). This finding highlights the importance of using a well-trained initial model as the foundation for the SASWISE approach.
\begin{figure}[!t]
\centering
\captionof{table}{Quantitative metrics for Bayesian models. Presented as mean with standard deviation (in parentheses). We use the Wilcoxon signed-rank test against the baseline single U-Net model among the testing dataset, with the Bonferroni-corrected significance level alpha = 0.05 represented by bolded numbers. MAE = mean absolute error, RMSE = root mean squared error, SSIM = structural similarity index measure, PSNR = peak signal-to-noise ratio. Acutance is defined as the average of the absolute gradient values, which are derived using the Sobel operator.}
\label{table_ReSyn}
\includegraphics[width=\linewidth]{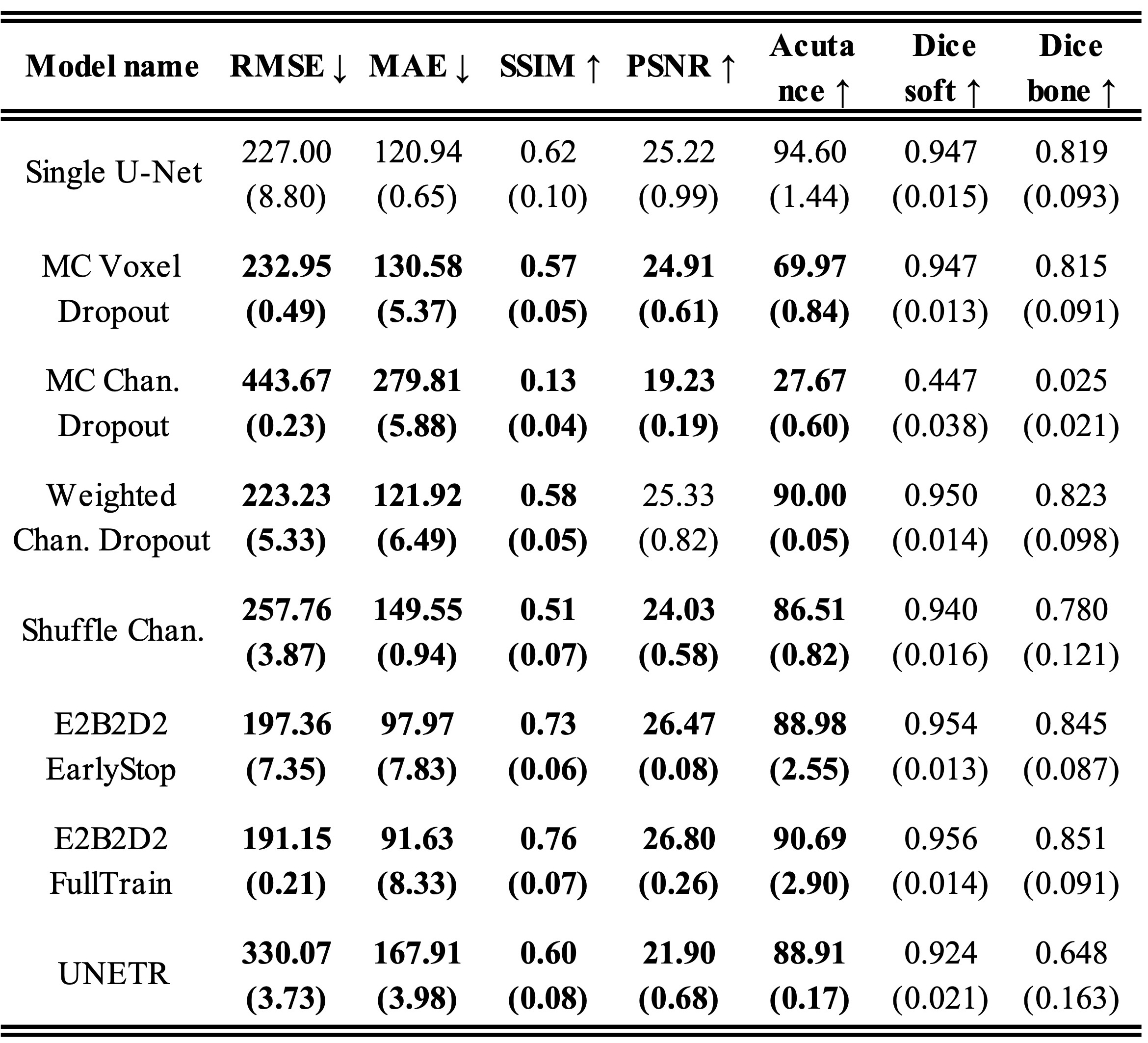}
\end{figure}
\subsection{Pruning}
\begin{figure*}[!t]
\centering
\includegraphics[width=0.85\linewidth]{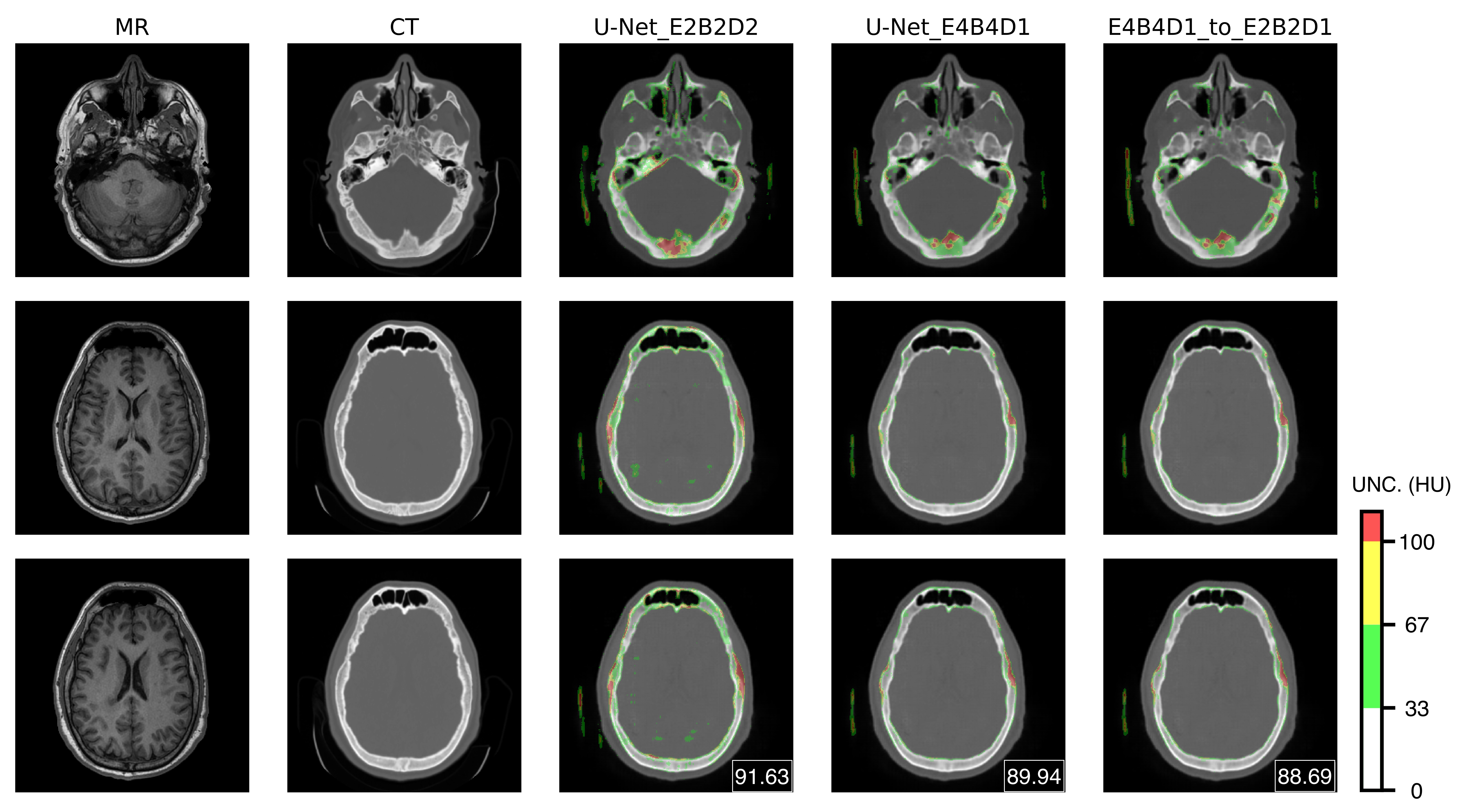}
\caption{The images showcase the effects of model pruning on MR-to-CT image synthesis across three axial sections. The third column illustrates synthesis results from a U-Net model configured with two encoder/bottleneck/decoder blocks, while the fourth column displays results from a model equipped with four such blocks. The fifth column features images from a model initially trained with four blocks, then pruned to two blocks. Accuracy in the synthesized images is indicated by color overlays: transparency signifies standard deviations below 33 Hounsfield Units, green for 33-67 HU, yellow for 67-100 HU, and red for deviations exceeding 100 HU. The mean absolute error (MAE) is noted in the bottom right in Hounsfield Units.}
\label{fig_RePrune}
\end{figure*}
Table 3 displays the results of the model pruning process in the synthesis approach. The pruned models (E1B4D4, E4B4D1, E4B4D4) outperform the base U-Net, illustrating that they maintain high performance while reducing complexity. This emphasizes the potential of model pruning for boosting performance and computational efficiency by selecting fewer, high-performing blocks from the ensemble model. The improved results suggest that the pruning process effectively identifies optimal block combinations, contributing to overall enhanced performance.
\begin{figure}[!t]
\centering
\captionof{table}{Performance of E1B4D4, E4B4D1, and E4B4D4 models. Both full and distilled models are examined, showing mean and standard deviation (in parentheses). The Wilcoxon signed-rank test is used to compare full and distilled models against the baseline model, with alpha = 0.05 Bonferroni-corrected significance level denoted in bold. MAE = mean absolute error, RMSE = root mean squared error, SSIM = structural similarity index measure, PSNR = peak signal-to-noise ratio, Acutance is defined as the average of the absolute gradient values, which are derived using the Sobel operator.}
\label{table_RePrune}
\includegraphics[width=\linewidth]{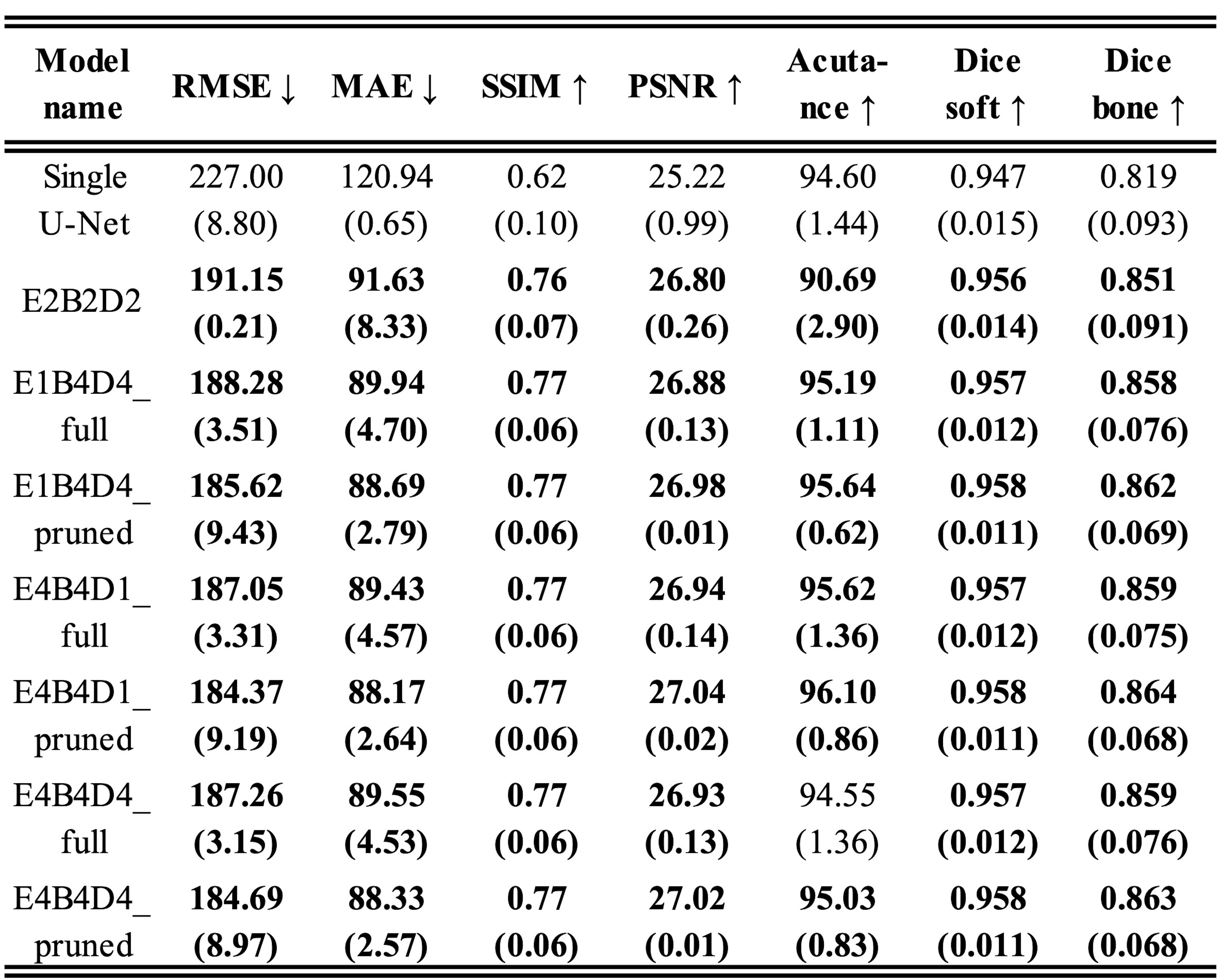}
\end{figure}
\subsection{Correlation analysis}
\begin{figure*}[!t]
\centering
\includegraphics[width=0.85\linewidth]{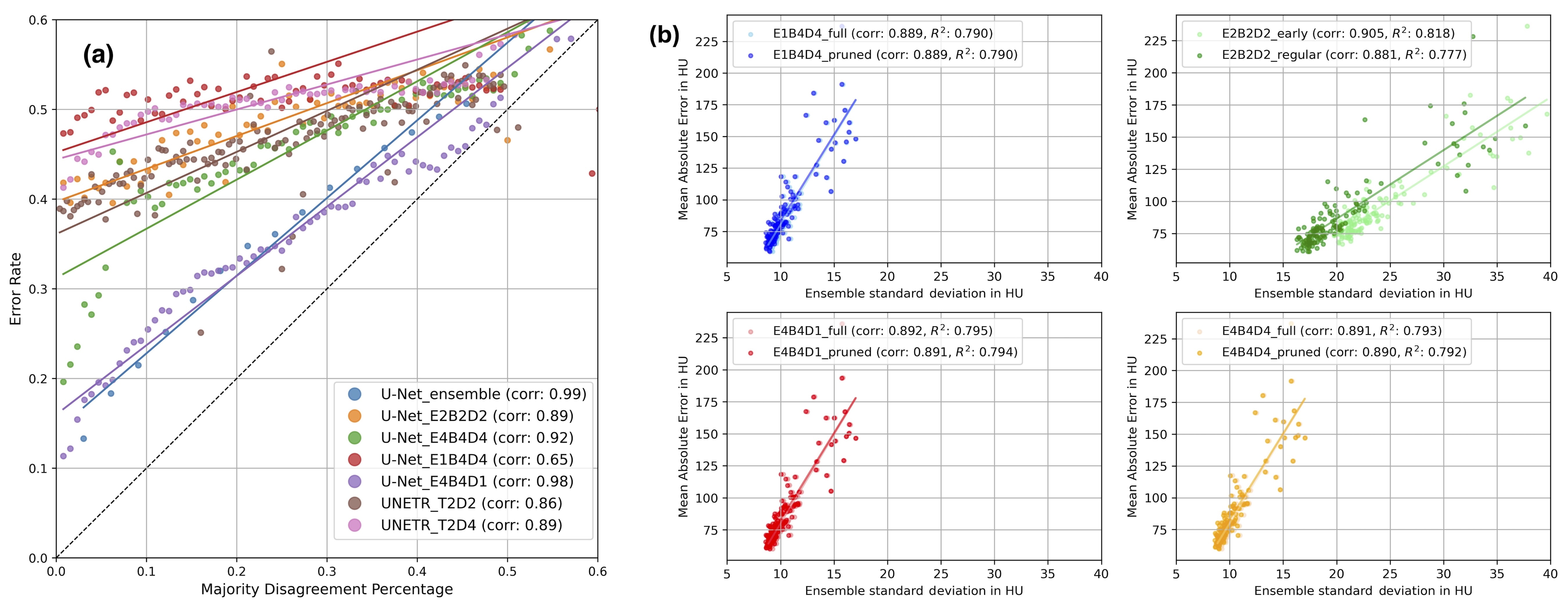}
\caption{(a) This graph presents an analytical view of voxel-level uncertainty in segmentation tasks, with the x-axis depicting deviation from the majority consensus and the y-axis representing the error rate. The correlation shown indicates that an increase in uncertainty is likely associated with reduced segmentation accuracy, especially at different levels of majority agreement. Pearson’s coefficients are computed for each scenario. (b) Displayed are four case-level correlation visualizations between the ensemble standard deviation and the final mean absolute error in Hounsfield Units (HU). Pearson's coefficients indicate that higher case mean uncertainty generally suggests higher mean errors. Each subplot also reveals that model pruning has minimal impact on this correlation, with regularly trained models performing better than those subjected to early stopping.}
\label{fig_Corre}
\end{figure*}
In the segmentation task, regions of high error typically occur near the boundaries, which correspond with areas of heightened uncertainty, as confirmed by the quantitative analysis shown in Figure \ref{fig_Corre}(a). In the synthesis task, a visible trend suggests that uncertainty is correlated with error, as illustrated in Figure \ref{fig_Corre}(b). While model pruning does not enhance this correlation, starting from a fully converged model before initiating diversification can enhance the performance of the final ensemble results.
\begin{figure}[!t]
\centering
\captionof{table}{This table presents the Dice and IoU metrics for various block configurations and training statuses of a model, categorized by mask type (low and high). Each entry shows the mean and standard deviation (in parentheses) of the respective metric. Models are grouped by block setting (E2B2D2, E1B4D4, E4B4D1, E4B4D4) and training status (EarlyStop, Regular, Full, Pruned), providing a comparative overview of performance across different architectural adjustments and training conditions.}
\label{table_RePrune}
\includegraphics[width=\linewidth]{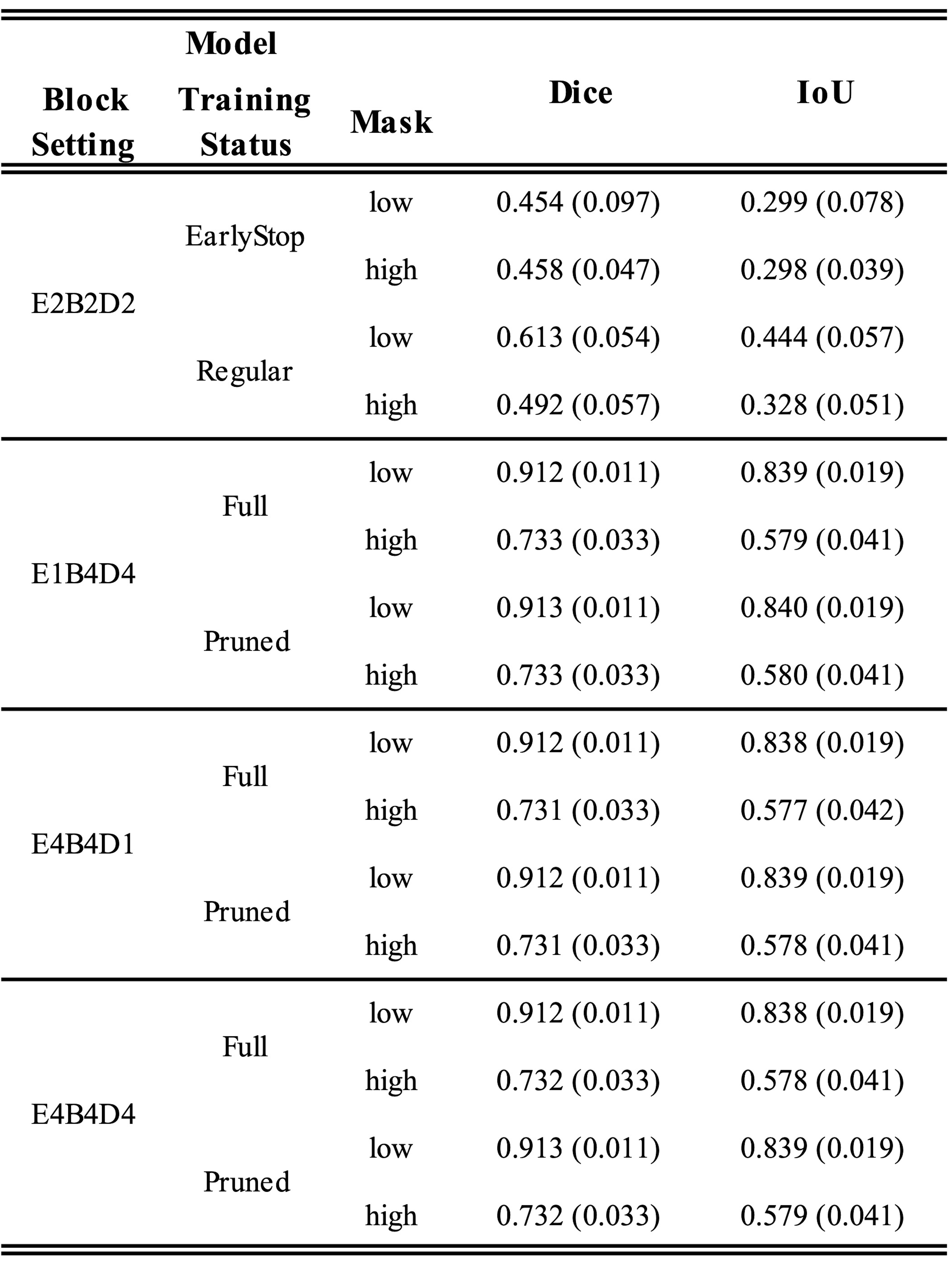}
\end{figure}
\begin{figure*}[!t]
\centering
\includegraphics[width=0.85\linewidth]{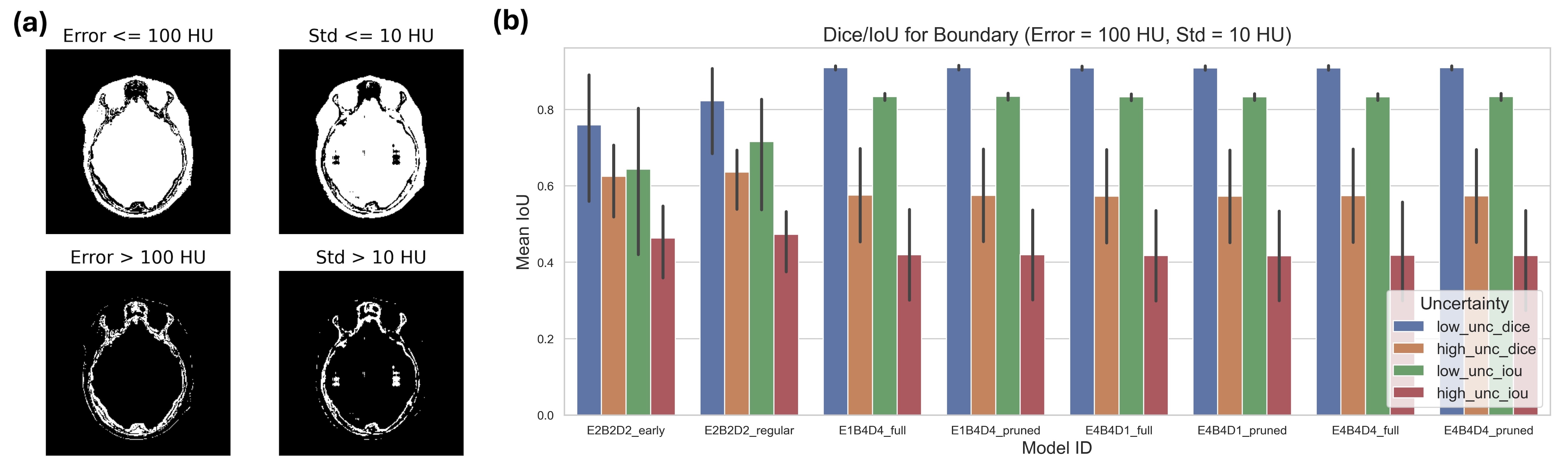}
\caption{(a) Displays masks indicating regions of higher/lower error and higher/lower uncertainty, illustrating significant overlap between these areas. (b) Presents quantitative results for Dice coefficients and IoU scores, comparing the boundaries for error (100 Hounsfield Units) and uncertainty (10 Hounsfield Units) across various models.}
\label{fig_diceIoU}
\end{figure*}
For the overlapping analysis shown in Figure \ref{fig_diceIoU}, it is observed that regions of low error and uncertainty overlap significantly, as do the regions of high error and high uncertainty. The numerical data support this observation. This indicates that uncertainty mapping can be effectively used to identify regions likely to have high error rates.
\subsection{Corruption}
\begin{figure*}[!t]
\centering
\includegraphics[width=0.85\linewidth]{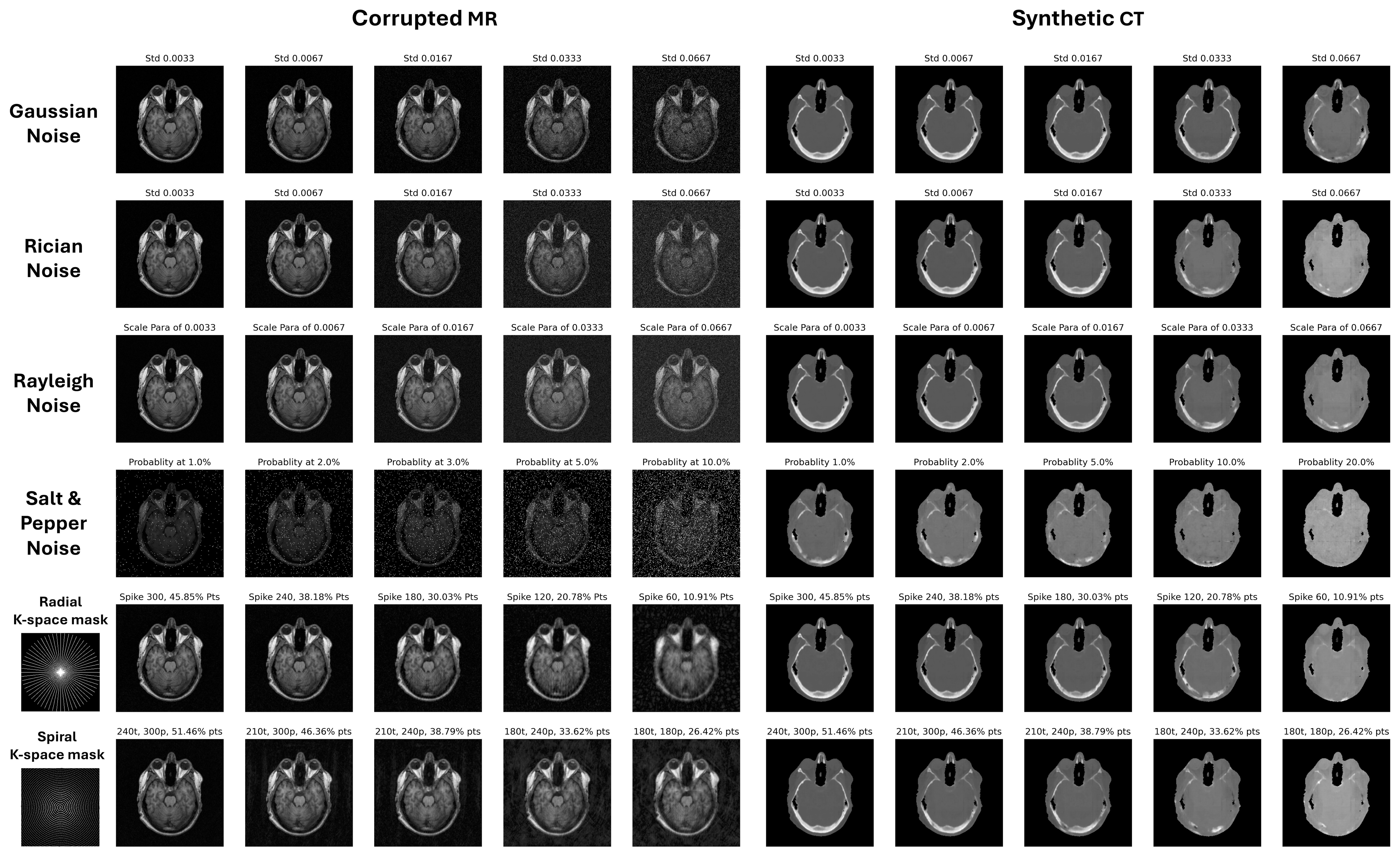}
\caption{This visualization demonstrates the impact of various corruption types on image quality, with increasing parameters leading to deteriorated image clarity. For Gaussian and Rician noise, the corruption level is dictated by the standard deviation of the Gaussian distribution. Rayleigh noise adjustment is based on the scale factor, while Salt \& Pepper noise depends on the probability that a pixel will be set to its minimum or maximum value. For Radial K-space masks, the parameters include the number of radial spikes per axial slice and the percentage of sampling points compared to regular sampling. For Spiral K-space masks, parameters consist of the number of turns (circles) and points per turn, alongside the comparative number of sampling points per axial slice. These undersampling techniques are depicted with K-space masks on the far left, each labeled beneath its respective mask name. The right half of the image showcases the synthetic CT images predicted from these corrupted MR inputs, illustrating the practical effects of each corruption type.}
\label{fig_MRICorru}
\end{figure*}
Six types of corrupted MR images are displayed in Figure \ref{fig_MRICorru}, along with the corresponding synthetic CT images reconstructed from these corrupted inputs. It is observed that synthetic CT images are prone to failure at relatively high levels of corruption. Specifically, the use of Salt \& Pepper noise proves to be unrealistic for MR-to-CT synthesis. Additionally, with K-space undersampling, while the MR images may appear normal with fewer sampling points, the synthetic CT images start to show signs of failure, indicating a decrease in synthesis quality under such conditions. These observations are further examined in Figure \ref{fig_CorreCorru}, where it is noted that correlation coefficients remain high at low levels of corruption but diminish as corruption levels increase. Notably, all levels involving Salt \& Pepper noise consistently fail, underscoring that this type of noise is particularly problematic and difficult to manage effectively.
\begin{figure*}[!t]
\centering
\includegraphics[width=0.85\linewidth]{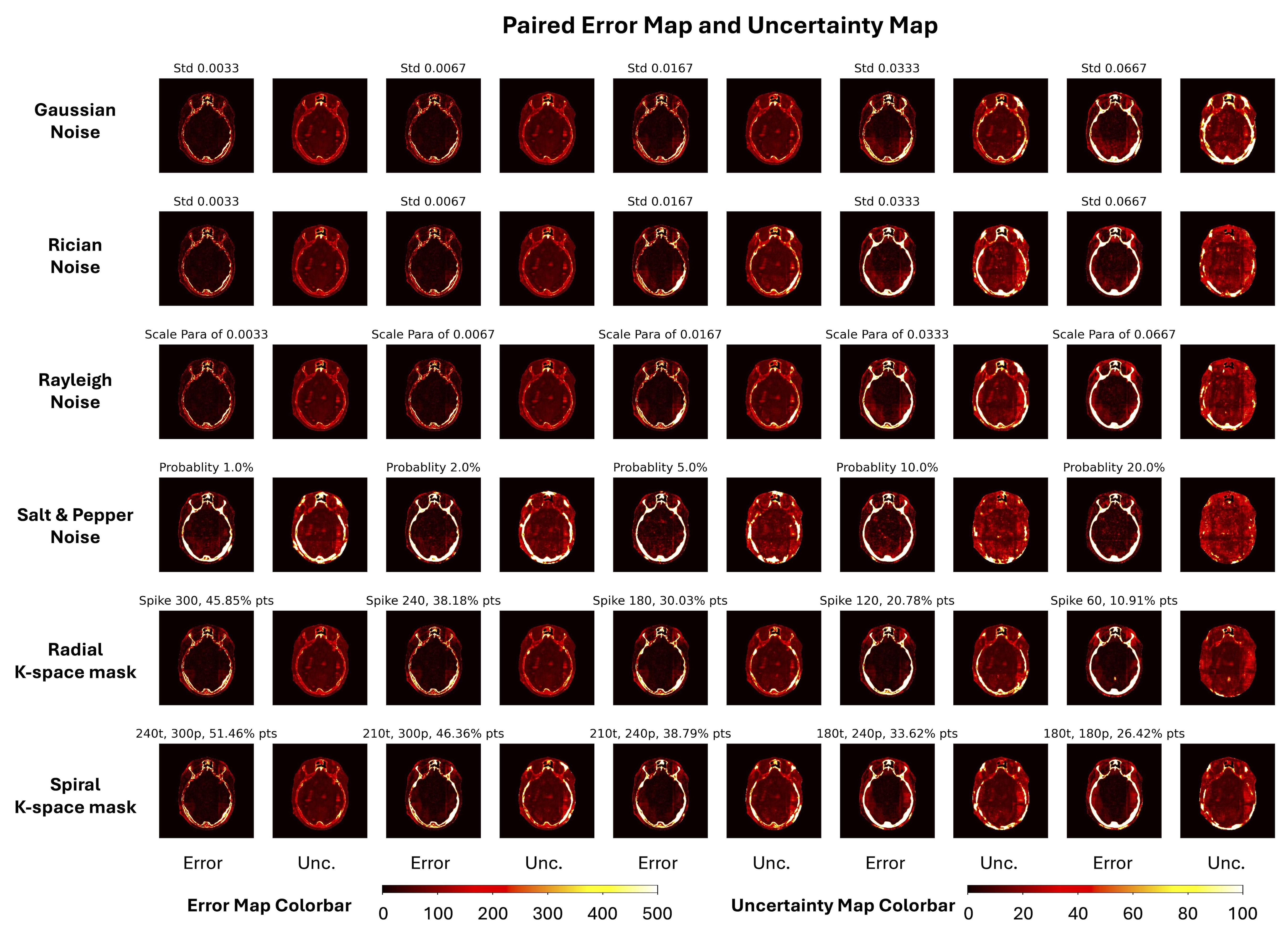}
\caption{These figures illustrate the correlation between uncertainty and error, displayed side by side. At low levels of corruption, correlations are still evident. However, as the corruption level increases, the effectiveness of the uncertainty map begins to diminish, particularly failing at high levels of corruption.}
\label{fig_ReCorru}
\end{figure*}
\begin{figure*}[!t]
\centering
\includegraphics[width=0.85\linewidth]{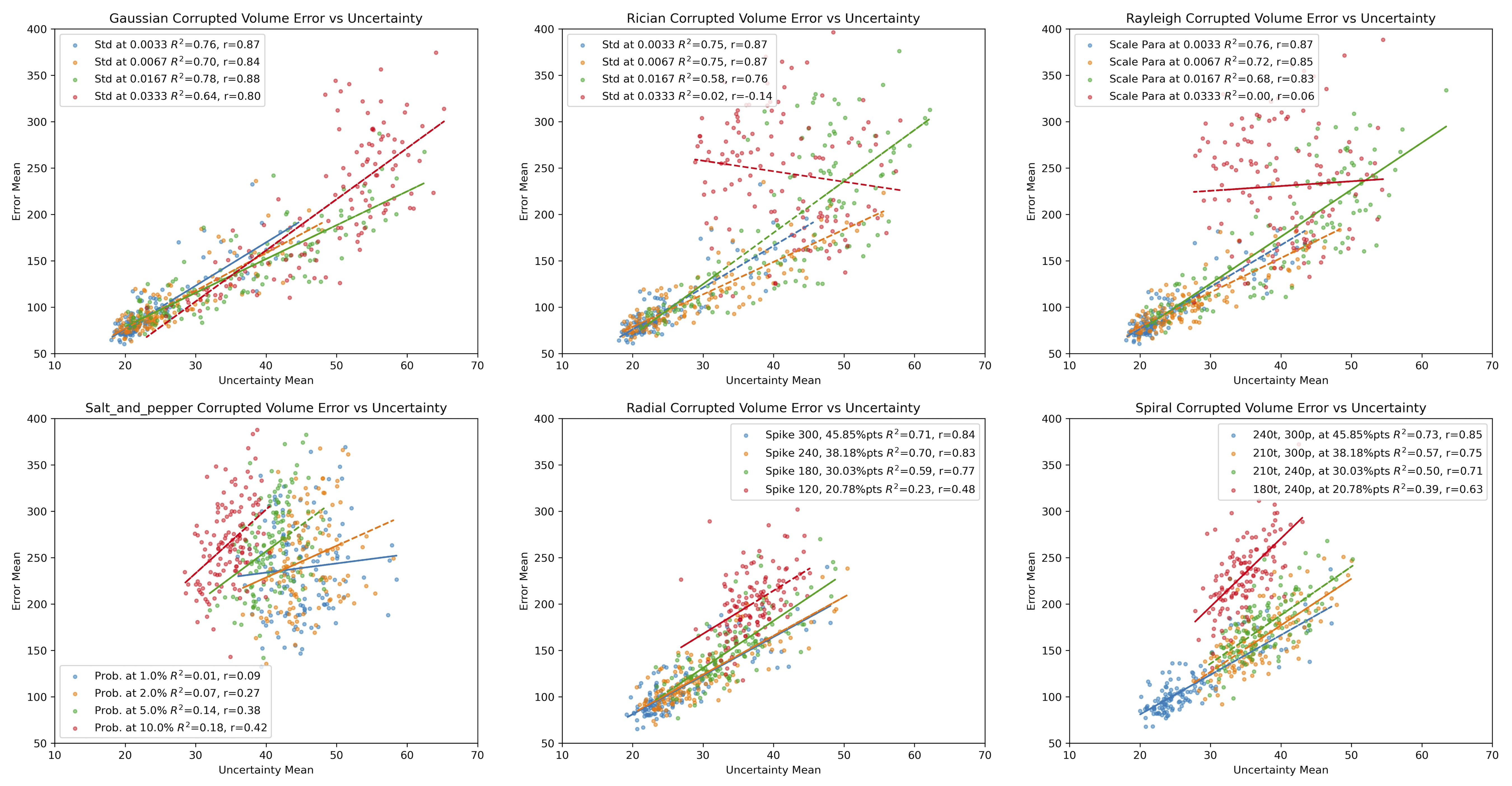}
\caption{Quantitative correlation analysis for corruption. In each corruption type, correlations are analyzed in case level for various corruption levels, and the Pearson's coefficients ($r$)and Coefficient of determination ($R^2$) are also computed.}
\label{fig_CorreCorru}
\end{figure*}
\begin{figure*}[!t]
\captionof{table}{This table displays the effects of various corruption methods on MRI image quality, showing prediction errors, standard deviations (in parentheses), and correlation metrics (Pearson's r and R-squared values) across different levels of each noise type. The corruption methods include Gaussian, Rician, Rayleigh, Salt \& Pepper noises, and Radial and Spiral K-space masks, each affecting the image quality and accuracy differently. The table succinctly captures how each corruption level impacts the synthetic outcomes and predictive reliability.}
\label{table_ReCoru}
\centering
\includegraphics[width=\linewidth]{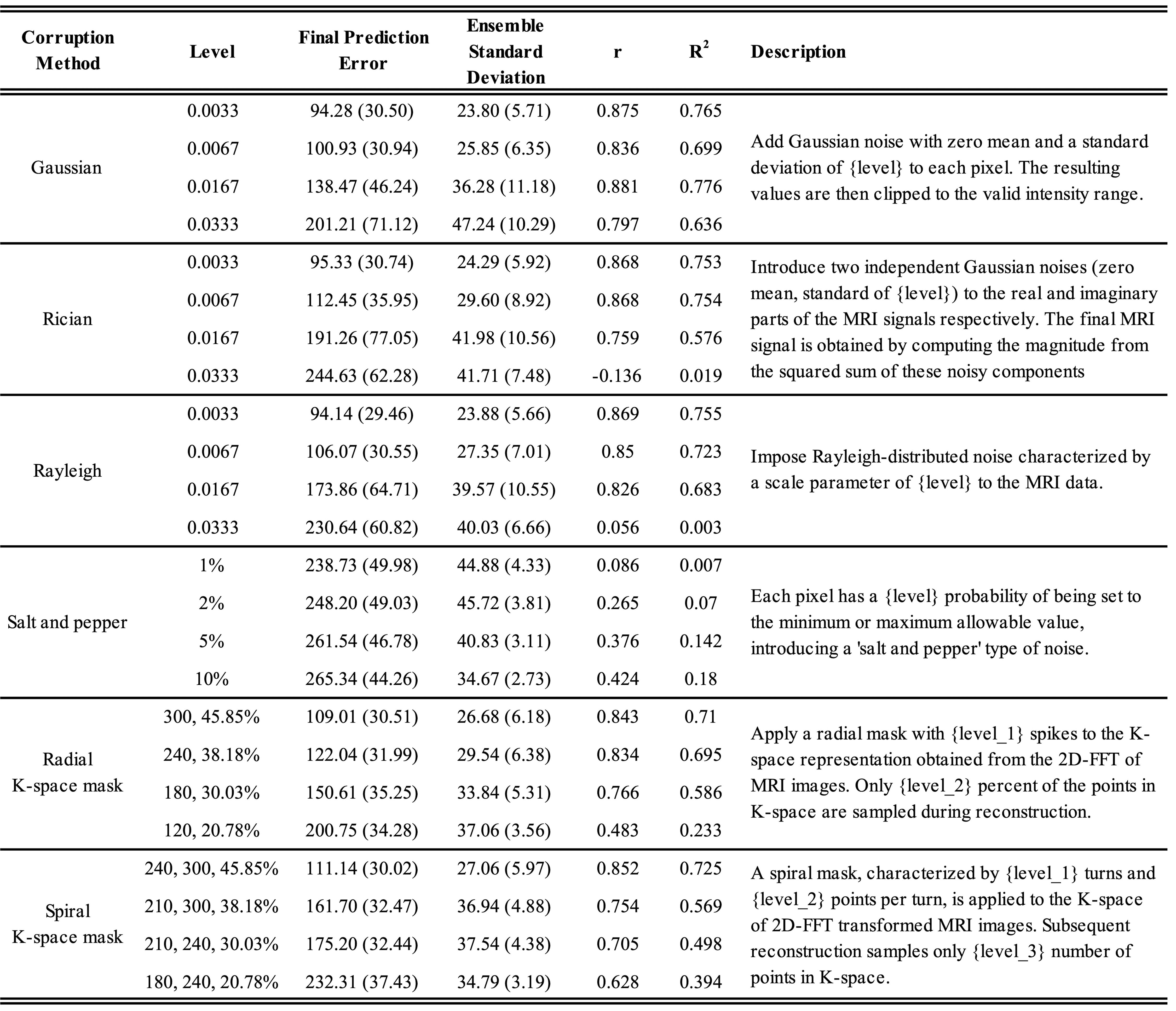}
\end{figure*}
\section{Discussion}
\subsection{Efficient training and resource utilization}
The SASWISE system enhances ensemble modeling by reconstructing models at the sub-model level, thus necessitating fewer effective epochs and resources compared to traditional ensemble methods. While conventional ensemble models, such as those discussed in (\cite{huangSnapshotEnsemblesTrain2017}), capture $N$ model snapshots using 
$O(N)$ computations, SASWISE employs block-level reconstitution, requiring only $O(\log N)$ computational resources. Specifically, if a model comprises $P$ blocks with $M$ candidate blocks each, the potential maximum number of models in the final output pool is 
$M^P$. This structure not only allows for rigorous downstream statistical analysis but also promotes diversity among model checkpoints. This increased efficiency facilitates the creation of larger and more diverse ensembles, which enhance performance without overburdening computational resources. It's important to note that the evaluation of ensembles, whether conventional or SASWISE, demands an equivalent amount of computational resources.

Furthermore, the SASWISE technique does not necessitate additional GPU memory. Since only one effective model is executed at a time, extra copies can be transferred to the CPU during training to minimize GPU usage. This approach ensures that the SASWISE system can be implemented with the same GPU resources as conventional models, at the cost of extended training time due to CPU-GPU transferring I/O time. This efficient resource utilization makes the SASWISE method a more scalable and practical alternative for creating ensemble models.
\subsection{Minimal modifications to existing models}
The SASWISE system distinguishes itself from other uncertainty estimation methods by employing a unique approach that differs significantly from techniques used in other studies. Unlike methods such as those cited in (\cite{araujoDRGRADUATEUncertaintyaware2020}), which add an additional layer, (\cite{nikovskiConstructingBayesianNetworks2000}) that incorporate an extra decoder, (\cite{sensoyEvidentialDeepLearning2018}) that refactor the loss, (\cite{lakshminarayananSimpleScalablePredictive2017}) that adjust hyper-parameters, (\cite{karimiAccurateRobustDeep2019}) that augment the training dataset, or (\cite{leeEfficientEnsembleModel2020}) that use different random seeds, the SASWISE method avoids these modifications which can introduce unpredictable effects on model performance.

Instead, SASWISE leverages optimal training practices during the regular training phase. In the diversification stage, it strategically diverges all models from the most effective pre-established checkpoints. This method ensures that each model in the ensemble starts from a high-quality baseline, minimizing variations in performance due to initial conditions, model components, loss formulations, hyperparameters, or training data differences. This structured approach in SASWISE not only simplifies the ensemble creation process but also maintains consistent performance across different models, ensuring robustness and reliability in uncertainty estimation.
\subsection{Practical implications and clinical relevance}
In our methodology, uncertainty serves as a measure of divergence among models or a deviation from the benchmark checkpoint. It reveals the extent of consensus and identifies regions with notable divergence. High uncertainty areas may necessitate expert review or human intervention. This approach highlights the importance of initiating from a well-trained baseline, which allows for an exploration around the pre-trained checkpoint and a deep understanding of the local minima landscape. By providing optimized checkpoints within this region, our strategy ensures a more robust and reliable ensemble, thereby enhancing decision-making across various applications.

In the context of segmentation tasks, as depicted in Figure \ref{fig_ReSeg}, medical professionals can leverage the displayed uncertainty to efficiently review and potentially refine the segmentation boundaries. This functionality facilitates quick identification of areas where segmentation may have been missed or requires manual correction. In synthesis tasks, as illustrated in Figures \ref{fig_ReSyn} and \ref{fig_RePrune}, areas highlighted in red indicate transitions between different tissue types (such as air to soft tissue or soft tissue to bone), pointing to possible inaccuracies or artificially introduced micro-structures in the synthesized image. The significant overlap between error and uncertainty masks, as shown in Figure \ref{fig_diceIoU}, corroborates this observation. Furthermore, the strong correlation between case-level error and uncertainty, demonstrated in Figure \ref{fig_Corre}, suggests that the system can prompt alerts for human review when abnormal uncertainty is detected in individual cases. This correlation persists even with increasing levels of potential corruption in the input images, demonstrated in Figure \ref{fig_CorreCorru}.


Compared to various dropout implementations, ranging from node-level (\cite{galDropoutBayesianApproximation2015}), through channel-level (\cite{houWeightedChannelDropout2019}), to layer-level (\cite{leeEfficientEnsembleModel2020}), dropout is fundamentally designed to prevent complex co-adaptations among nodes, enhancing generalizability and mitigating model overfitting (\cite{DBLP:journals/corr/abs-1207-0580}). However, in image synthesis, dropout may impede neural networks from learning sharp, distinct features by forcing them to rely on safer, more general representations across different input combinations, rather than specific, aggressive features consistently paired with the same inputs. This limitation suggests a hypothesis: a block, or a functional cohort of layers, could act as a projection function within the representation space. In the diversification stage, similar but distinct candidate blocks could project the same input into proximate areas within the representation space. This cascading diversification allows for the exploration of a neighborhood around a well-established checkpoint input, revealing the variance inherent in the model's predictions. Such an approach not only broadens the interpretative capacity of the model but also enhances its adaptability by detailing the range of potential outputs given slight variations in input handling.
\subsection{Limitations and future plans}
This study acknowledges several limitations that warrant further exploration. First, the segmentation task utilized a limited dataset; expanding its size and diversity, although resource-intensive, could facilitate a more comprehensive performance evaluation. Second, exploring interdependencies and co-adaptations between candidate blocks could optimize ensemble construction. Third, the current fusion and uncertainty functions are basic; employing more sophisticated alternatives could enhance results and refine the interpretation of model certainty. Lastly, expanding the investigation into the block-level ensemble framework’s broader applications, use cases, and advanced statistical analyses could further validate SASWISE's utility and strengthen its robustness.

Additionally, this sub-model combination could potentially support a federated learning framework, mitigating certainty influence from individual participants. For instance, if each candidate block is trained by a different agent, blocks can be discarded if an agent is either poisoning the model or wishes to withdraw. Transferring sub-model blocks, rather than the entire model, could reduce communication costs and bolster data privacy since it is challenging to reconstruct training samples from only parts of the model.

This approach might also serve as an active learning framework. Given that expert medical segmentation is costly, we could initiate training with a small dataset, employing the SASWISE framework to estimate uncertainty on the remainder of the unlabeled dataset. Leveraging the correlation between uncertainty and error allows for the identification of challenging data samples, potentially reducing the burden on medical experts required to label the dataset.  
\section{Conclusion}
This paper introduces an efficient sub-model ensemble framework that establishes a model pool with exponentially numerous models, starting from a well-trained baseline. This framework was evaluated through image segmentation and synthesis tasks, demonstrating enhanced performance from the initial point. A correlation analysis between error and estimated uncertainty was conducted, revealing a strong correlation that facilitates human attention guidance and model failure detection without the need for ground truth. The study also investigated the impact of various input corruptions, showing that the correlation persists under medium-level corruption and undersampling. With minimal modifications to well-established models and no additional hardware resources required, this proposed framework holds significant potential for advancing uncertainty estimation in the medical imaging field.

\section*{CRediT authorship contribution statement}
Weijie Chen: Conceptualization, Methodology, Software, Validation, Formal analysis, Investigation, Writing - Review \& Editing, Visualization.

Alan McMillan: Methodology, Software, Validation, Formal analysis, Investigation, Resources, Data Curation, Writing - Review \& Editing, Visualization, Supervision, Project administration, Funding acquisition.

\section*{Declaration on generative AI}
Statement: During the preparation of this work the author(s) used ChatGPT in order to improve readability and enhance fluency. After using this tool/service, the author(s) reviewed and edited the content as needed and take(s) full responsibility for the content of the publication.



\section*{Acknowledgement}
We thank Iman Zare Estakhraji for his contributions in data collection and pre-processing. This work was supported by the National Institutes of Health R01EB026708 and 5R01LM013151.

\bibliographystyle{model2-names.bst}\biboptions{authoryear}
\bibliography{SASWISE_BibLaTeX.bib}

\end{document}